%% file: neurips_2026.tex
\PassOptionsToPackage{numbers, compress}{natbib}
\documentclass{article}
\usepackage[main,final]{neurips_2026}

\usepackage[most]{tcolorbox}
\usepackage{xcolor}
\usepackage{enumitem}

\usepackage{booktabs}
\usepackage{tabularx}
\usepackage{array}

\usepackage{multirow}
\usepackage{graphicx}
\usepackage[table]{xcolor}

\usepackage[utf8]{inputenc} 
\usepackage[T1]{fontenc}    
\usepackage{url}            
\usepackage{amsfonts}       
\usepackage{nicefrac}       
\usepackage{microtype}  
\usepackage{wrapfig}
\usepackage{multicol}
\usepackage{multirow}
\usepackage{booktabs}
\usepackage{amsmath,amsfonts}
\usepackage{amssymb}
\usepackage{algorithmic}
\usepackage{algorithm}
\usepackage{graphicx}
\usepackage{enumitem}
\usepackage{xspace}
\usepackage{caption}
\usepackage{pifont}
\usepackage{subcaption}
\usepackage{natbib}
\usepackage{colortbl}
\usepackage{xcolor}         
\usepackage[colorlinks,
            linkcolor=red,
            anchorcolor=blue,
            citecolor=cyan
            ]{hyperref}
\newcommand{\ie}{\emph{i.e., }}
\newcommand{\eg}{\emph{e.g., }}

\newcommand{\cf}{\emph{cf. }}

\usepackage[most]{tcolorbox}
\usepackage[table]{xcolor}
\usepackage{enumitem}

\definecolor{softred}{HTML}{FDECEC}
\definecolor{softgreen}{HTML}{FFF7E6}
\definecolor{softblue}{HTML}{EEF5FF}
\definecolor{darkred}{HTML}{B91C1C}
\definecolor{darkgreen}{HTML}{B7791F}
\definecolor{darkblue}{HTML}{1D4ED8}

\newtcolorbox{ourscasebox}[1][]{
  enhanced,
  breakable,
  colback=softgreen,
  colframe=darkgreen!55,
  boxrule=0.7pt,
  arc=2pt,
  left=6pt,
  right=6pt,
  top=5pt,
  bottom=5pt,
  fonttitle=\bfseries,
  coltitle=black,
  #1
}

\newtcolorbox{sampleinputbox}[1][]{
  enhanced,
  breakable,
  colback=gray!4,
  colframe=gray!45,
  boxrule=0.6pt,
  arc=2pt,
  left=6pt,
  right=6pt,
  top=6pt,
  bottom=6pt,
  fonttitle=\bfseries,
  coltitle=black,
  #1
}

\newtcolorbox{samplelabelbox}[1][]{
  enhanced,
  breakable,
  colback=cyan!3,
  colframe=cyan!35!gray,
  boxrule=0.6pt,
  arc=2pt,
  left=6pt,
  right=6pt,
  top=6pt,
  bottom=6pt,
  fonttitle=\bfseries,
  coltitle=black,
  #1
}

\newtcolorbox{systempromptbox}[1][]{
  enhanced,
  breakable,
  colback=gray!5,
  colframe=gray!65,
  boxrule=0.6pt,
  arc=2pt,
  left=6pt,
  right=6pt,
  top=6pt,
  bottom=6pt,
  fonttitle=\bfseries,
  coltitle=black,
  #1
}

\newtcolorbox{userpromptbox}[1][]{
  enhanced,
  breakable,
  colback=blue!3,
  colframe=blue!35,
  boxrule=0.6pt,
  arc=2pt,
  left=6pt,
  right=6pt,
  top=6pt,
  bottom=6pt,
  fonttitle=\bfseries,
  coltitle=black,
  #1
}

\newtcolorbox{casepromptbox}[1][]{
  enhanced,
  breakable,
  colback=softblue,
  colframe=darkblue!55,
  boxrule=0.7pt,
  arc=2pt,
  left=6pt,
  right=6pt,
  top=6pt,
  bottom=6pt,
  fonttitle=\bfseries,
  coltitle=black,
  #1
}

\newtcolorbox{baselinebox}[1][]{
  enhanced,
  breakable,
  colback=softred,
  colframe=darkred!55,
  boxrule=0.6pt,
  arc=2pt,
  left=6pt,
  right=6pt,
  top=5pt,
  bottom=5pt,
  fonttitle=\bfseries,
  coltitle=black,
  #1
}

\usepackage[table]{xcolor}
\definecolor{ourscyan}{RGB}{220,245,248}

\title{Teaching Large Language Models When Not to Know: Learning Temporal Critique for Ex-Ante Reasoning}

\author{%
\textbf{Chenlu Ding}$^{1}$ \quad
\textbf{Jiancan Wu}$^{1,}$\thanks{Corresponding authors.} \quad
\textbf{Yanchen Luo}$^{2}$ \\ \quad
\textbf{Zheyuan Liu}$^{3}$ \quad
\textbf{Yancheng Yuan}$^{2,*}$ \quad
\textbf{Xiang Wang}$^{1}$ \\[1mm]
$^{1}$University of Science and Technology of China \\
$^{2}$The Hong Kong Polytechnic University \\
$^{3}$University of Notre Dame \\
\texttt{dingchenlu200103@gmail.com} 
}

\begin{document}

\maketitle
\input{chapter/0_abs}

\input{chapter/1_intro_new}

\input{chapter/2_preliminary}
\input{chapter/3_method}
\input{chapter/4_exp}

\input{chapter/5_conclusion}

\bibliographystyle{unsrt}
\bibliography{nips}

\input{chapter/7_appendix}

\newpage
\input{checklist.tex}

\end{document}

%% file: chapter/0_abs.tex
\begin{abstract}
Large language models (LLMs) often fail to reason under temporal cutoffs: when prompted to answer from the standpoint of an earlier time, they exploit knowledge that became available only later. We study this failure through the lens of ex-ante reasoning, where a model must rely exclusively on information knowable before a cutoff. Through a systematic analysis of prompt-level interventions, we find that temporal leakage is highly sensitive to cutoff formulation and instruction placement: explicit cutoff statements outperform implicit historical framings, and prefix constraints reduce leakage more effectively than suffix constraints. These findings indicate that prompting can steer models into a temporal frame, but does not endow them with the ability to verify whether a response is temporally admissible. We further argue that supervised fine-tuning is insufficient, since ex-ante correctness is not an intrinsic property of an answer, but a relation between the answer and the cutoff. To address this gap, we propose \textbf{TCFT}, a \underline{T}emporal \underline{C}ritique \underline{F}ine-\underline{T}uning framework that trains models to acquire cutoff-aware temporal verification. Given a query, a cutoff, and a candidate response, TCFT teaches the model to identify post-cutoff leakage, explain temporal boundary violations, and judge temporal admissibility. Experiments with Qwen2.5-7B-Instruct and Qwen2.5-14B-Instruct show that TCFT consistently outperforms prompting and SFT baselines, reducing average leakage by 41.89 and 37.79 percentage points\footnote{Our code is available at \url{https://github.com/oceanoceanna/TCFT_2026}.}, respectively.
\end{abstract}

%% file: chapter/1_intro_new.tex
\section{Introduction}
\textit{Suppose today is May 1, 2015. How would you describe Twitter?}
A human answering this question would naturally inhabit that moment --- recalling a 140-character platform led by Dick Costolo --- while instinctively setting aside what came later: Musk's 2022 acquisition, and the rebranding to X.
This capability is referred to as \textbf{ex-ante reasoning} \cite{ExAnte}, the ability to reason under a strict temporal cutoff, drawing exclusively on information available up to that point.
Far from a curiosity, it underpins any application whose validity depends on \textit{what was knowable at the time}, from backtesting investment strategies \cite{FinLLMs,Accounting} and conducting historical counterfactual analysis \cite{history} to tracing the genuine progression of scientific ideas \cite{science}.
In each, value hinges not on what a model can retrieve, but on \textbf{what it refuses to use}.
Yet, today's most capable language models \cite{GPT-4,Gemini} routinely fail at this refusal \cite{ExAnte}:
ask the question above with the cutoff stated unambiguously, they volunteer post-2015 facts about Musk and X, and this failure does not diminish with scale.
Stronger models memorize post-cutoff knowledge more thoroughly, and thus leak it more fluently (\cf Figure \ref{fig:evolve}) \cite{ExAnte,Dated_Data}.

Confronted with such leakage, the natural first recourse is to push harder on the prompt: state the cutoff more forcefully, elicit chain-of-thought \cite{cot}, or invoke self-verification \cite{self-verification}.
We systematically investigate this family of interventions, treating their behavior as a diagnostic probe into how models handle temporal constraints.
Two findings stand out, each difficult to reconcile with the hypothesis that models possess any stable notion of a cutoff:
\begin{itemize}[leftmargin=*]
\vspace{-7pt}
    \item \textbf{Leakage tracks surface phrasing, not semantics.} Two logically equivalent cutoff specifications --- an implicit historical framing (\textit{``Assume today is May 1, 2009''}) and an explicit knowledge restriction (\textit{``Use only information available before May 1, 2009''}) --- yield markedly different leakage rates, with explicit phrasings consistently outperforming implicit ones.
    \vspace{-3pt}
    \item \textbf{Leakage tracks position, not proximity to the answer.} Placing the cutoff before the query (prefix) reduces leakage substantially more than placing it afterward (suffix), despite the latter's closer proximity to generation.
    \vspace{-5pt}
\end{itemize}
Taken together, these findings reveal an underlying weakness: prompt-based interventions modulate \textit{how the model is steered into a temporal frame}, but they do not install a mechanism that asks, of each candidate fact, whether it is admissible under the given cutoff.
\textbf{In essence, the model is being conditioned, not verified.}
This diagnosis also explains why supervised fine-tuning offers no real escape: SFT is an imitation objective \cite{CritiqueFineTuning,lora}, teaching the model to reproduce a target answer for a given prompt \cite{instrcut,ft,sft,Self-Instruct}. But ex-ante correctness is not a property of an answer in isolation. Instead, it is a relation between an answer and a cutoff.
No volume of answer-imitation supervision can instill this contingency, because the supervision signal itself never asks the model to evaluate the relation.

What is missing, then, is supervision that operates on the relation rather than on the answer --- training that teaches the model not what to say, but why a given response is or is not temporally admissible at a given time.
We instantiate this principle as \textbf{TCFT} (\underline{T}emporal \underline{C}ritique \underline{F}ine-\underline{T}uning), a critique-based \cite{CritiqueFineTuning} training framework for ex-ante reasoning.
Rather than fine-tuning the model to imitate temporally correct answers, TCFT trains it to acquire cutoff-aware temporal verification.
Given a query, a cutoff, and a candidate response, the model learns to assess whether the response is temporally admissible, identify post-cutoff leakage, and explain boundary violations.
In this way, TCFT directly supervises the missing relation between an answer and a cutoff, rather than treating ex-ante reasoning as a matter of prompt conditioning or answer imitation.
We instantiate this supervision by using a teacher model to generate cutoff-conditioned critiques for candidate responses, where each critique highlights temporally invalid facts and provides an admissibility judgment \cite{Verify_step,judgement}.
The student model is then fine-tuned on these critiques, acquiring the ability to verify whether a candidate response is valid under the specified cutoff.
With this temporal verification capability, the model can reject temporally inadmissible facts before accepting an answer, thereby reducing the leakage of future knowledge.
Experiments with Qwen2.5-7B-Instruct and Qwen2.5-14B-Instruct \cite{qwen-2.5} demonstrate that TCFT consistently outperforms strong prompting and fine-tuning baselines, achieving better adherence to temporal constraints with leakage rate reductions of 41.89 and 37.79 percentage points, respectively.
\begin{figure}[t]
\vspace{-10pt}
    \centering  
    \includegraphics[width=1.0\textwidth]{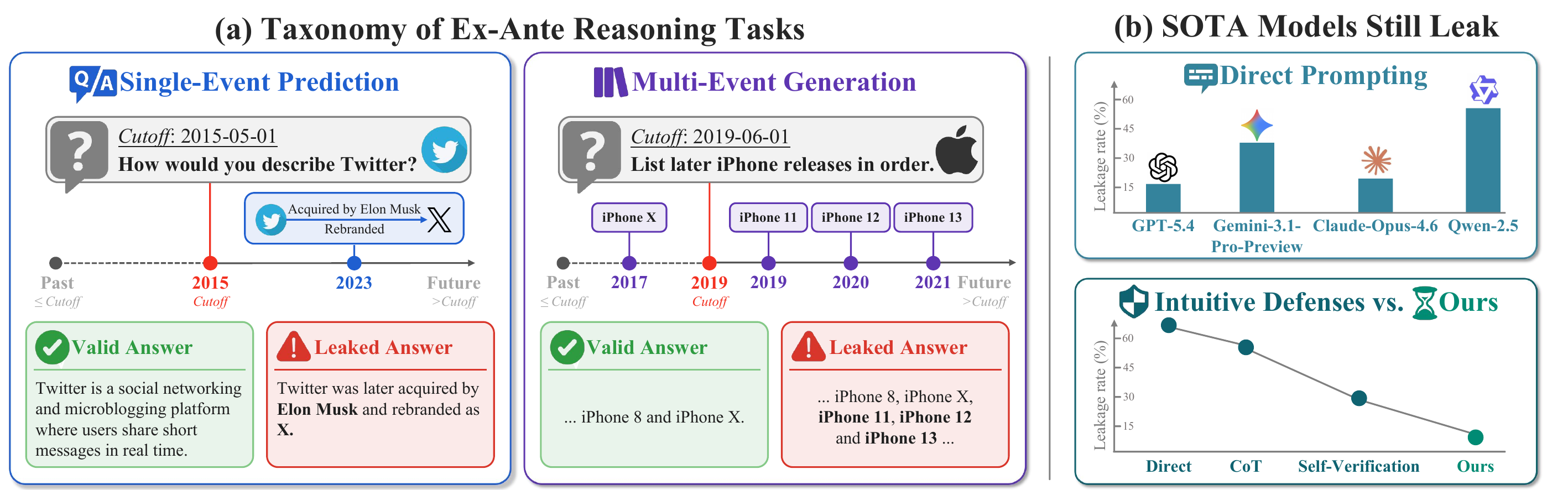}
    \caption{Overview of ex-ante reasoning tasks, temporal leakage in SOTA models, and the comparison between prompting-based defenses and our TCFT.}
    \label{fig:intro}
    \vspace{-15pt}
\end{figure}

%% file: chapter/2_preliminary.tex
\section{Preliminary} \label{sec:preliminary}

To ground our empirical investigation and subsequent methodology, we first formalize the ex-ante reasoning problem, outline the task taxonomy, and establish rigorous evaluation metrics.

\subsection{Task Definition}
Given a query $x$ and a specific cutoff timestamp $t_c$, ex-ante reasoning aims to produce a response $\hat{y}$ using only information verifiable before $t_c$. 
Let $\mathcal{A}(x)$ denote the universe of all factual knowledge associated with query $x$.
We define a temporal mapping $\tau(f)$ that assigns each fact $f \in \mathcal{A}(x)$ the exact time it became verifiable in the real world.
The temporal constraint $t_c$ naturally partitions this knowledge space into two disjoint subsets:
\begin{align}
    R_{\text{prev}}(x, t_c) &= \{f \in \mathcal{A}(x) \mid \tau(f) \leq t_c\} \\
    R_{\text{post}}(x, t_c) &= \{f \in \mathcal{A}(x) \mid \tau(f) > t_c\}
\end{align}
where $R_{\text{prev}}(x, t_c)$ represents the admissible historical context, and $R_{\text{post}}(x, t_c)$ denotes future facts that must remain inaccessible.  For a model $M$ parameterized by $\theta$, the output is $\hat{y} = M(x, t_c)$.
An oracle ideal response $y^*$ is constructed entirely from $R_{\text{prev}}(x, t_c)$ and contains only the pre-cutoff information. Strict ex-ante validity therefore requires excluding all future knowledge: $F(\hat{y}) \cap R_{\text{post}}(x,t_c) = \varnothing$. Depending on the structural complexity of the required response, the ex-ante reasoning task can be categorized into the following two primary paradigms~\cite{ExAnte}:
\begin{itemize}[leftmargin=*]
\vspace{-5pt}
\item \textbf{Single-Event Prediction.}
Single-event prediction asks the model to adopt the perspective of $t_c$ and predict an outcome verifiable only afterward. This paradigm typically appears in time-sensitive QA or numerical forecasting.
For an event whose true outcome $a^*$ materializes at $t_e > t_c$, ex-ante reasoning requires a historically grounded prediction or abstention, rather than leaking the future ground truth.
\vspace{-3pt}
\item \textbf{Multi-Event Generation.}
In open-ended generation, such as Wikipedia-style \cite{Wikipedia} summarization from a historical perspective, the response $\hat{y}$ is a complex text sequence rather than a single fact.
For evaluation, we decompose it into independent atomic claims 
$F(\hat{y}) = \{f_1, f_2, \ldots, f_n\}$, 
each of which is verified against the cutoff $t_c$.
\vspace{-5pt}
\end{itemize}

\subsection{Evaluation Metrics}

The core failure mode in ex-ante reasoning is \textit{temporal leakage}: the unintended use of post-cutoff knowledge ($R_{\text{post}}$) under an explicit pre-cutoff restriction ($R_{\text{prev}}$). Leakage differs from \textit{memorization}: the latter refers to the model's internalization of facts that became verifiable only after $t_c$, which is a prerequisite for leakage rather than a failure in itself.
Without prior memorization of the relevant future facts, cutoff adherence is trivial; leakage evaluation is therefore meaningful only when such memorization is established. We formalize leakage by measuring whether the generated output intersects with the restricted future knowledge space. 

For \textbf{single-event prediction}, leakage is a binary indicator of whether the model directly reveals the future outcome:
\begin{equation}
    L_{\text{single}}(x) = \mathbf{1}[\hat{y} = a^* \land t_e > t_c].
\end{equation}
This captures direct disclosure of a fact that should be unknown at cutoff $t_c$. In the more complex setting of \textbf{multi-event generation}, we calculate a claim-level leakage rate, defined as the proportion of generated atomic facts that fall into the restricted future knowledge set:
\begin{equation}
    L_{\text{multi}}(x) = \frac{1}{n}\sum_{i=1}^{n}\mathbf{1}[f_i \in R_{\mathrm{post}}(x,t_c)].
\end{equation}
A query is flagged as leaked if any generated claim depends on post-cutoff information. A high leakage rate indicates a systemic failure of temporal reasoning: the model cannot reliably separate pre-cutoff knowledge from post-cutoff facts.
Unlike forecasting~\cite{gruver2023large,jin2023time}, ex-ante reasoning concerns the \textit{admissibility} of information.
Thus, even a factually ``correct'' future prediction is a failure if it relies on post-cutoff information.

Evaluating temporal adherence requires capturing both the violation of constraints and the utility of the generated text. Relying solely on constraint satisfaction is insufficient, as models might exploit degenerate strategies to bypass the rules. This joint criterion prevents temporal gains from coming at the cost of task utility. The validity evaluation metric is detailed in Appendix~\ref{app:validity_metric}.

\section{Investigating the Brittleness of Prompt-based Control} \label{sec:insight}

The most intuitive approach to enforcing temporal boundaries in LLMs is through prompt engineering.
However, prior evaluations suggest that simple instructions often fail to consistently eliminate temporal leakage~\cite{ExAnte}.
To better understand the limitations of prompt-based interventions, we systematically investigate how instruction paradigms affect model adherence to temporal cutoffs.

\subsection{Formalizing Temporal Instruction Paradigms}
To systematically analyze the impact of prompting, we represent a temporal template $T$ as a tuple $(t_c, e_T, \phi_T, \ell_T)$, capturing key dimensions of prompt-based temporal control. Here, $t_c$ denotes the target cutoff date and $e_T \in \{\text{implicit}, \text{explicit}\}$ specifies boundary explicitness.
The predicate $\phi_T(f_i; t_c) \in \{0, 1\}$ evaluates whether a fact $f_i$ is admissible under template $T$ and cutoff $t_c$.
Finally, $\ell_T \in \{\text{prefix}, \text{suffix}\}$ indicates whether the temporal constraint is placed before or after the query.

Given this formalization, we investigate how variations in $e_T$ and $\ell_T$ affect ex-ante reasoning performance, focusing on two dimensions: the explicitness of the temporal constraint and its placement within the prompt.
These factors are hypothesized to influence how effectively a model can internalize and adhere to temporal boundaries, thereby affecting leakage rates and overall validity.

\subsection{Insight 1: Leakage Tracks Surface Phrasing}
The first critical finding concerns how the phrasing of temporal boundaries influences model behavior.
In ex-ante reasoning, models are typically prompted into a temporal frame either implicitly through historical framing (\eg ``Assume today is May 1, 2009'') or explicitly through direct knowledge restriction (\eg ``Use only information available before May 1, 2009'').

Formally, given a candidate fact $f_i$, a template induces an admissibility predicate $\phi_T(f_i; t_c) \in \{0, 1\}$.
A strong template explicitly verbalizes this predicate, ideally requiring $\forall f_i \in \mathcal{F}(y), \phi_T(f_i; t_c) = 1$, whereas a weak template only implicitly points to the same boundary.
Specifically:
\begin{itemize}[leftmargin=*]
\vspace{-5pt}
    \item \textbf{Explicit templates} specify the knowledge boundary, often through prohibitive language (\eg \textit{``Use only information available before $t_c$''} or \textit{``Do not rely on facts after $t_c$''}). We find that these templates consistently reduce leakage across both single-event and multi-event tasks. By verbalizing $\phi_T(f_i; t_c) = \mathbf{1}[\tau(f_i) \le t_c]$, explicit prompts give the model a clear surface rule to follow.
    \vspace{-3pt}
    \item \textbf{Implicit templates} instead frame the task through a historical persona or perspective (\eg \textit{``Assume today is $t_c$''} or \textit{``Imagine you are answering on $t_c$''}). Although the temporal boundary is identical, models perform substantially worse under implicit constraints. This sensitivity suggests that current LLMs rely on surface linguistic cues rather than a stable, cutoff-aware reasoning mechanism.
    \vspace{-5pt}
\end{itemize}

As shown in Figure~\ref{fig:insight}, explicit templates consistently yield lower future-information leakage than implicit templates.
This formulation sensitivity suggests that current models lack a stable understanding of temporal boundaries. If models had mastered cutoff-aware reasoning, surface-level phrasing of the same boundary should not substantially change performance~\cite{ExAnte}. Instead, models appear to be linguistically conditioned rather than temporally verified.

\subsection{Insight 2: Leakage Tracks Prompt Position}
The second finding involves the positional variation of constraints. Beyond formulation, the position $\ell_T$ of the temporal constraint also affects adherence: it can appear either as a prefix before the query or as a suffix after it. For instance, consider the following two templates for a query about Twitter:
\begin{itemize}[leftmargin=*]
\vspace{-5pt}
    \item \textbf{Prefix}: ``Use only information available before May 1, 2009. How would you describe Twitter?''
    \vspace{-3pt}
    \item \textbf{Suffix}: ``How would you describe Twitter? Use only information available before May 1, 2009.''
    \vspace{-5pt}
\end{itemize}
This positional variation is orthogonal to formulation explicitness, allowing us to independently assess how placement affects model behavior.

\begin{figure*}[t]
    \centering
    \vspace{-7pt}
    \begin{subfigure}[t]{0.418\linewidth}
        \centering
\includegraphics[width=\linewidth]{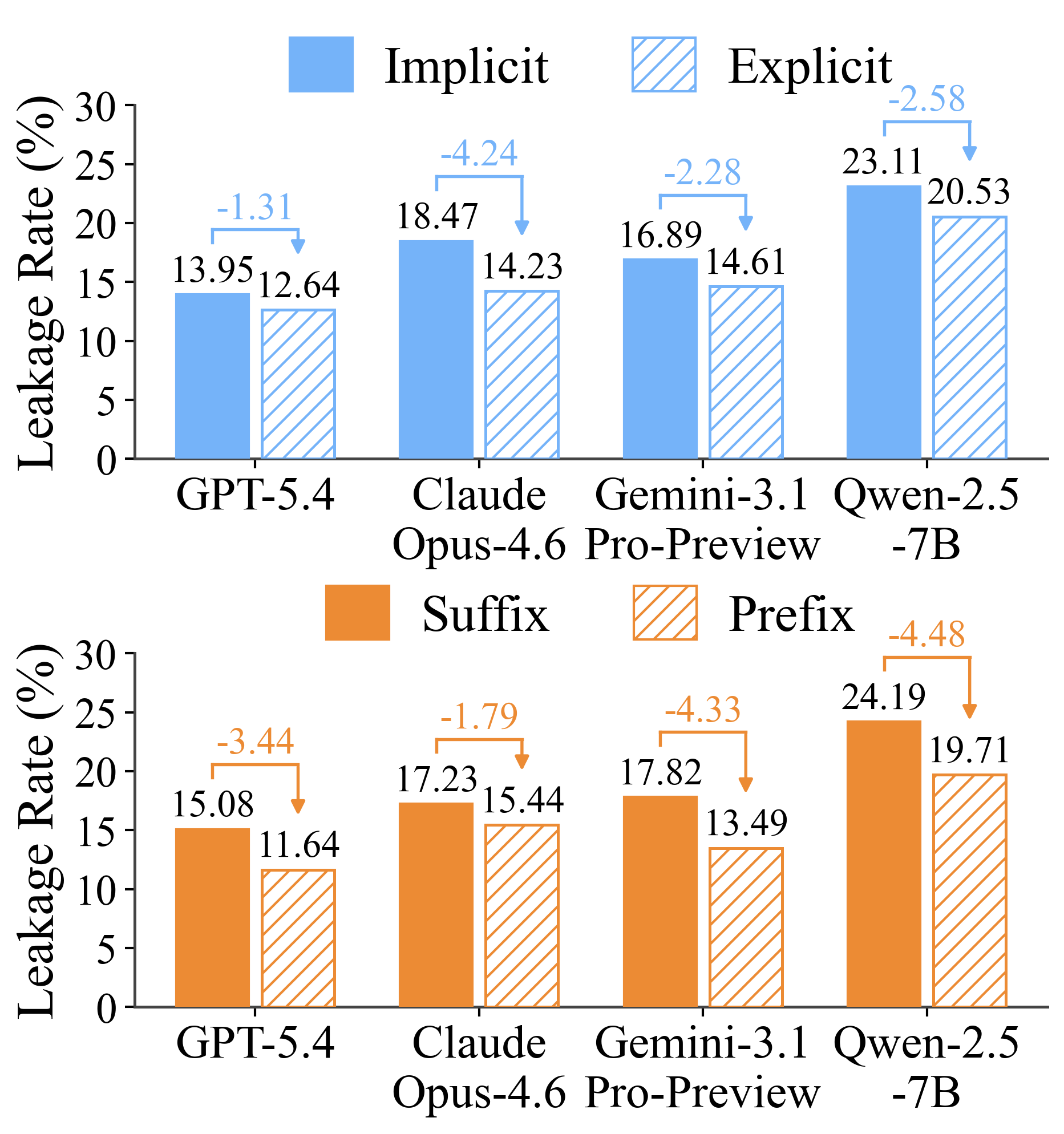}
        \caption{Temporal template sensitivity.}
        \label{fig:insight}
    \end{subfigure}
    \hfill
    \begin{subfigure}[t]{0.5225\linewidth}
        \centering
\includegraphics[width=\linewidth]{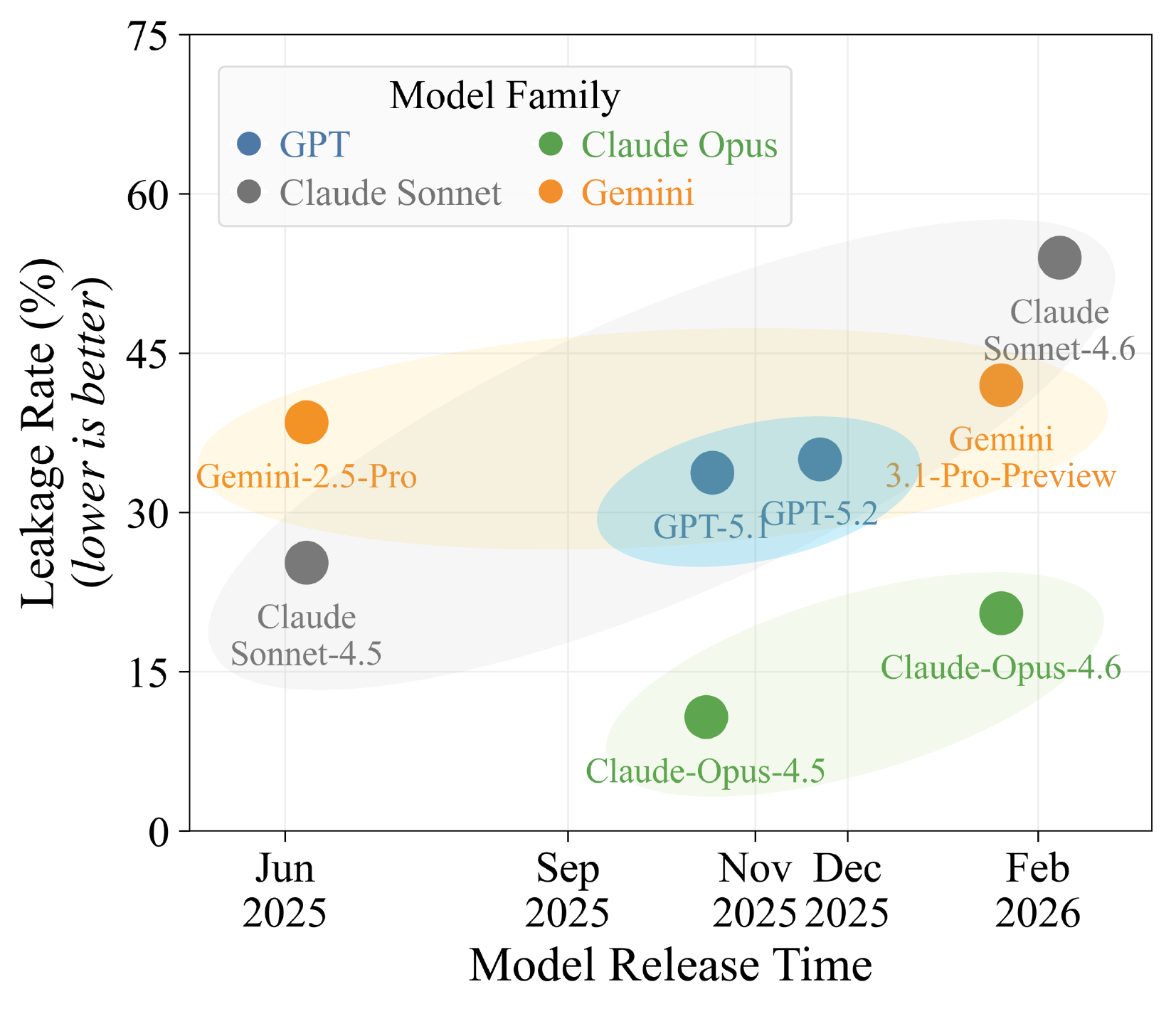}
        \caption{QA leakage of proprietary LLMs over release time.}
        \label{fig:evolve}
    \end{subfigure}
    \caption{Investigating the brittleness of prompt-based control and temporal leakage in proprietary LLMs on the QA task. Effect of temporal constraint formulation and placement on QA leakage (\ref{fig:insight}). Model capability does not guarantee cutoff awareness (\ref{fig:evolve}).}
    \label{fig_distribution}
    \vspace{-10pt}
\end{figure*}

Empirically, we observe a clear \textbf{prefix advantage}: constraints placed at the beginning of the prompt yield markedly lower leakage than those appended as a suffix, as shown in Figure~\ref{fig:insight}.
This finding is counterintuitive, since suffix constraints are closer to generation and might be expected to exert stronger control.
Instead, the result aligns with known \textit{positional biases} in in-context learning and long-context processing~\cite{OrderMatters,gruver2023large}, where information near the start of a prompt is often more stable.
For ex-ante reasoning, the prefix advantage suggests that temporal boundaries function as global conditioning signals that are more easily ignored when treated as trailing instructions.
Consequently, prompt engineering induces position-sensitive conditioning rather than robust temporal verification.

%% file: chapter/3_method.tex
\section{Method}\label{sec:method}
We address the core challenge identified in Section~\ref{sec:insight}: ex-ante reasoning is fundamentally a temporal verification problem rather than a generation problem.
Prompt-based control and conventional SFT fall short because they separate answer production from valid-at-cutoff judgment.
Instead of imitating target answers, our approach trains models to critique candidate responses under temporal constraints.
The model learns to assess whether a response is admissible at cutoff $t_c$, identify temporal violations, and explain its judgment.
By making temporal validity a direct training target, this framework enables more robust cutoff-aware reasoning across diverse cutoff specifications.

\subsection{Rethinking the Learning Objective: From Imitation to Verification} \label{sec:method_motivation}
Standard supervised fine-tuning optimizes $\min_\theta \mathbb{E}_{(x,y^*)} \ell(M_\theta(x), y^*)$, where $y^*$ is a target answer.
This treats correctness as an answer-level property independent of context.
In ex-ante reasoning, however, validity is cutoff-dependent: the validity of $y$ depends on the temporal boundary $t_c$:
\begin{equation}
	\text{Valid}(y \mid x, t_c) \neq \text{Valid}(y \mid x).
\end{equation}
Thus, answer-matching supervision alone cannot teach a model when not to use memorized post-cutoff facts.
Section~\ref{sec:insight} reveals that this limitation runs deeper than prompt phrasing. 

The formulation and positional sensitivity of temporal constraints suggest that models lack an internalized procedure for judging temporal admissibility. Prompt engineering only steers surface behavior, without installing a verification mechanism. We therefore reframe training to directly supervise cutoff-conditioned temporal judgments, shifting the objective from matching target answers to determining whether a response is valid under the given cutoff.

\subsection{Temporal Critique Task Definition} \label{sec:method_task}

In our TCFT framework, we extend the Critique Fine-Tuning (CFT) \cite{CritiqueFineTuning} paradigm to the temporal setting.
Classical CFT mainly targets reasoning verification: given a question and a potentially flawed solution, the model learns to critique the reasoning process and identify errors.
In our setting, however, critique is intrinsically time-dependent, since the validity of a claim depends on the temporal boundary.

Given a query $x$, cutoff timestamp $t_c$, and candidate response $y$, the teacher model $\mathcal{C}_\Theta$ (\eg GPT-5.4 \cite{OpenAI2026GPT54}) outputs a structured tuple:
\begin{equation}
	\mathcal{C}_\Theta(x, t_c, y) = (r, v),
\end{equation}
where $v$ is a binary verdict indicating whether $y$ is temporally admissible at cutoff $t_c$ and $r$ provides the supporting rationale.

This decomposition captures two coupled capabilities: \textbf{leakage detection}, which identifies claims in $y$ that are not verifiable at $t_c$; and \textbf{temporal reasoning}, which explains why those claims violate the cutoff based on their verifiability timelines.

\subsection{Temporal Critique Dataset Construction} \label{sec:method_data}
To construct temporal critique supervision, we pair each query with multiple cutoffs, candidate responses, and corresponding temporal verdicts.
The resulting dataset is structured as:
\begin{equation}
	\mathcal{D}_{\text{tc}} = \{(x_i, t_{c,i}, y_i, c_i)\}_{i=1}^{|\mathcal{D}|},
\end{equation}
where each instance $(x_i, t_{c,i})$ is paired with a critique-conclusion target $c_i = (r_i,v_i)$. We augment the data with diverse prompt formulations to improve cutoff-aware judgment in a task-specific manner:
\vspace{-5pt}
\paragraph{QA Augmentation.}
For each QA pair from the unaugmented ex-ante reasoning dataset \cite{ExAnte}, we generate variants along three dimensions:
(1) \textbf{Cutoff variation}, where the cutoff date is set to 1, 14, and 30 days before the event, exposing the model to the same factual context under different temporal boundaries;
(2) \textbf{Formulation variation}, where each cutoff is paired with explicit knowledge restriction (``Use only information available before $t_c$'') and implicit historical framing (``Assume today is $t_c$'');
and (3) \textbf{Position variation}, where each template is placed either before the query as a prefix or after it as a suffix. This results in $3 \times 2 \times 2 = 12$ distinct training instances per original pair.
\vspace{-5pt}
\paragraph{Wikipedia Augmentation.}
For generation tasks, we focus on formulation and position variation:
(1) cutoff dates are kept fixed, since Wikipedia-style prompts often involve historical entities or long-span events, making fine-grained cutoff shifts less well-defined; and (2) each example is expanded into $2 \times 2 = 4$ variants by crossing explicit/implicit formulation with prefix/suffix placement.

All splits (train, validation, and test) undergo the same augmentation, ensuring balanced exposure to strong and weak prompt formulations and enabling fair comparison across conditions.
This construction supervises temporal validity relations, \ie query-cutoff pairs with valid/invalid labels, rather than a single target answer.
By annotating cross-cutoff validity changes, the model learns that correctness is fundamentally contingent on $t_c$.

\subsection{Learning via Temporal Critique Supervision} \label{sec:method_objective}
In practice, TCFT is implemented as a supervised fine-tuning process over the temporal critique tuples.
Given a training instance $(x, t_c, y, c)$ with $c=(r,v)$, we optimize the following negative log-likelihood objective:
\begin{equation}
	\mathcal{L} = -\log p(c \mid x, t_c, y; \theta).
\end{equation}

The joint probability can be factorized as:
\begin{equation}
	p_\theta(c\mid x,t_c,y) = p_\theta(r\mid x,t_c,y)\cdot p_\theta(v\mid x,t_c,y,r),
\end{equation}
where the model first generates a rationale $r$ that examines the candidate response under the cutoff, and then summarizes this analysis into a verdict $v$.
This order encourages the verdict to be grounded in an explicit temporal critique rather than produced as an isolated label.
In implementation, the rationale and verdict are concatenated into a single critique-conclusion sequence and optimized via teacher forcing, yielding the full sequence likelihood.

Notably, this objective does not directly supervise final-answer generation (\eg predicting $y$ from $x$ and $t_c$).
Instead, the model learns to critique candidate answers and summarize the temporal judgment.
This inversion is deliberate: by decoupling generation from verification, the model can reuse its base knowledge while learning a conditional gate that flags leaked content.

\subsection{Inference} \label{sec:method_inference}
At inference time, the model directly generates a response $\hat{y}$ from the query and cutoff $(x,t_c)$.
Although TCFT is trained on temporal critiques rather than final answers, it equips the model with cutoff-aware verification.
By learning from temporally violated candidates and their critiques, the model internalizes which facts are admissible under a cutoff and which constitute leakage.
Thus, inference requires neither explicit critique generation nor answer refinement; the learned verification capability implicitly guides $\hat{y}$ away from post-cutoff information.

Formally, training optimizes the critique objective $p(c \mid x,t_c,y)$, while inference samples $\hat{y}$ from $p(\hat{y} \mid x,t_c)$.
Although the two stages differ in form, both rely on the same learned capability: judging temporal admissibility under the cutoff.
In this sense, TCFT does not teach the model to imitate a fixed target answer; instead, it teaches the model a temporal validity criterion that guides generation.

Accordingly, TCFT does not require an explicit answer-repair objective, \ie optimizing $p(y^* \mid x,t_c,y,c)$.
Rather than learning to rewrite a flawed answer into a target response, the model learns to recognize temporal admissibility itself, in line with prior critique-based learning~\cite{CritiqueFineTuning}.
Introducing a separate refinement objective would add a competing learning signal and dilute the focus on temporal verification.
Instead, the verdict $v$ and rationale $r$ provide direct supervision for temporal judgment, which can support downstream filtering or post-hoc correction when needed.

%% file: chapter/4_exp.tex
\section{Experiment}
This section provides a comprehensive experimental evaluation of TCFT, with the analysis structured around the following key research questions:
\textbf{RQ1:} How does TCFT perform compared with state-of-the-art closed-source models and strong baseline methods?
\textbf{RQ2:} How well does TCFT generalize across boundary explicitness $e_T$ and constraint placement $\ell_T$?
\textbf{RQ3:} How do the temporal critique rationale $r$ and verdict $v$ contribute to learning cutoff-aware verification?

\subsection{Experimental Setups}
We use Qwen2.5-7B-Instruct and Qwen2.5-14B-Instruct as the LLM backbones, excluding smaller models whose limited memorization makes leakage evaluation ill-defined. We focus on standard instruction models to isolate whether TCFT itself can induce cutoff-aware temporal verification, without relying on reasoning-specialized post-training.
We build on the original ex-ante reasoning benchmark, augmenting its single-event QA and Wikipedia-style generation tasks.
For evaluation, we use leakage rate as the primary metric: for QA, we report leakage rates under 1-day, 14-day, and 30-day cutoffs, together with the average leakage rate; for Wikipedia generation, we report leakage rate and claim accuracy.
We compare TCFT with closed-source models and open-source baselines, including Zero-shot, Few-shot, Chain-of-thought, Self-verification, and SFT.
Comprehensive details on the benchmark, baselines, evaluation metrics, and TCFT implementation are provided in Appendix~\ref{appendix:Hyperparameters}.

\begin{table*}[t]
\vspace{-5pt}
    \caption{Evaluation results on QA and Wiki. For QA, 1d, 14d, and 30d denote cutoff gaps of 1, 14, and 30 days before the target event date, and Avg. denotes the average QA leakage rate. For Wiki, we report leakage rate and claim accuracy. Lower leakage indicates stronger cutoff adherence, and higher accuracy indicates better factual utility.}
    \renewcommand{\arraystretch}{1.02}
    \centering
    \small
    \resizebox{0.85\textwidth}{!}{
        \input{table/qa_wiki_results}
    }
    \label{tab:main}
    \vspace{-5pt}
\end{table*}

\subsection{Results Analysis of Ex-Ante Reasoning (RQ1)}
We conduct experiments on two representative ex-ante inference tasks: QA and Wiki. Table~\ref{tab:main} reports results for proprietary LLMs and open-weight Qwen models under standard prompting, prompt-level interventions, SFT, and our method. Lower leakage indicates stronger cutoff adherence, and higher Wiki accuracy indicates better factual utility. We draw the following observations:
\begin{itemize}[leftmargin=*]
\vspace{-5pt}
    \item \textbf{Proprietary LLMs generally achieve stronger temporal adherence than open-weight baselines.} Among them, GPT-5.4 achieves the best overall performance, with 22.25\% QA leakage, 13.28\% Wiki leakage, and 81.37\% Wiki accuracy. This indicates that stronger general-purpose models exhibit better implicit temporal alignment, but still cannot reliably enforce strict cutoff constraints.
    \vspace{-2pt}
    \item \textbf{Prompt-level interventions do not consistently improve temporal adherence.} For the QA task, prompt-level interventions generally reduce leakage, with few-shot and self-verification reducing leakage by 52.80 and 50.93 percentage points, respectively, while CoT yields a smaller reduction of 14.18 percentage points.
    In contrast, prompt-level methods fail to consistently reduce leakage on the Wiki task. This suggests that prompt-level interventions mainly impose surface-level behavioral constraints, rather than enabling models to acquire genuine cutoff awareness.
    \vspace{-2pt}
    \item \textbf{Shorter cutoff gaps lead to higher QA leakage.}
    QA leakage is highest under the 1-day cutoff gap and decreases as the gap extends to 14 and 30 days, with average leakage rates of 40.00\%, 32.50\%, and 27.90\%, respectively. This observation is consistent with ExAnte~\cite{ExAnte}, suggesting that LLMs struggle most with near-cutoff events due to weak fine-grained temporal boundary awareness.
    \vspace{-2pt}
    \item \textbf{TCFT achieves the strongest temporal leakage control.}
    Compared with zero-shot, SFT reduces average QA and Wiki leakage by 45.15 and 4.03 percentage points, while TCFT achieves larger reductions of 70.00 and 9.68 percentage points, respectively. Meanwhile, TCFT improves Wiki accuracy from 65.58\% to 70.03\%. This suggests that the key to ex-ante reasoning is learning cutoff-aware verification, rather than simply imitating target answers.
    \vspace{-5pt}
\end{itemize}

\subsection{Generalization Across Temporal Templates (RQ2)}
\begin{figure}[htbp]
    \centering  
    \includegraphics[width=1.0\textwidth]{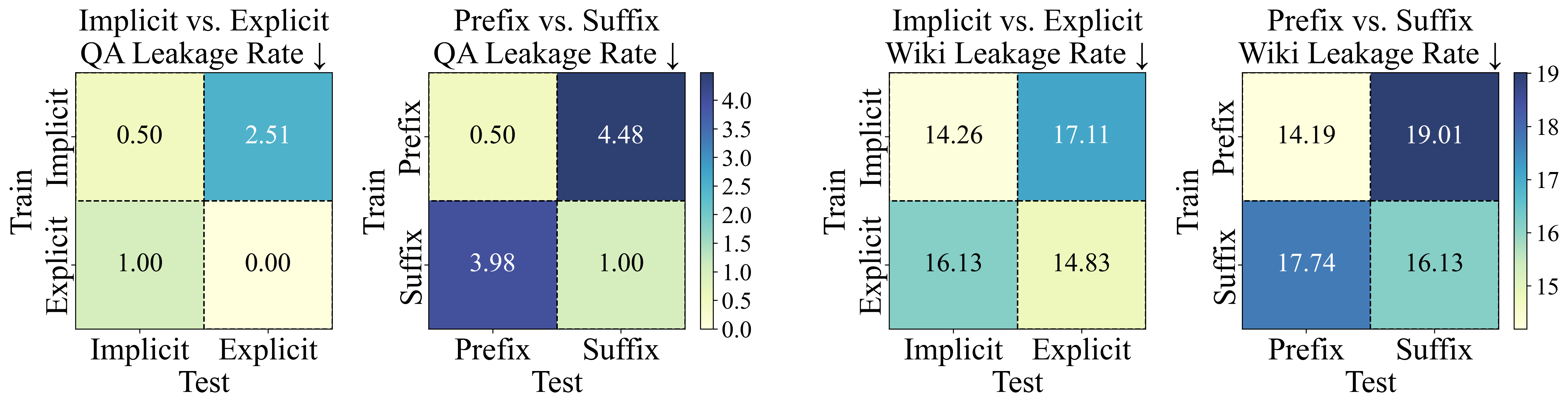}
    \caption{Cross-template generalization of TCFT on QA and Wiki with Qwen2.5-7B-Instruct. Rows denote the temporal template used for training, and columns denote the template used for testing.}
    \label{fig:cross}
\end{figure}

To investigate whether TCFT learns general temporal verification rather than template-specific patterns, we evaluate cross-template transfer on QA and Wiki. We vary templates along two dimensions: implicit vs. explicit temporal constraints and prefix vs. suffix cutoff placement. TCFT is trained on one template and tested on both the seen template and its counterpart, measuring template-specific performance and cross-template generalization. As shown in Figure~\ref{fig:cross}, we can find that:
\begin{itemize}[leftmargin=*]
    \item \textbf{TCFT remains effective across different temporal templates.}
    Across implicit/explicit formulations and prefix/suffix cutoff placements, TCFT consistently achieves low leakage. Under the same template setting, QA leakage remains within 0.00--1.00\%, and Wiki leakage ranges from 14.19\% to 16.13\%. These results suggest that TCFT captures the underlying cutoff-verification behavior across different surface formulations, rather than merely memorizing template-specific patterns.
    \item \textbf{TCFT remains effective under cross-template transfer.}
    When evaluated on unseen template variants, leakage increases only slightly and remains well controlled. Cross-template QA leakage stays low at 1.00--4.48\%, while Wiki leakage remains within 16.13--19.01\%. This suggests that TCFT acquires a transferable temporal verification capability rather than relying on prompt-specific surface patterns, enabling robust generalization across cutoff formulations.
\end{itemize}

\begin{figure}[htbp]
    \centering  
    \includegraphics[width=1.0\textwidth]{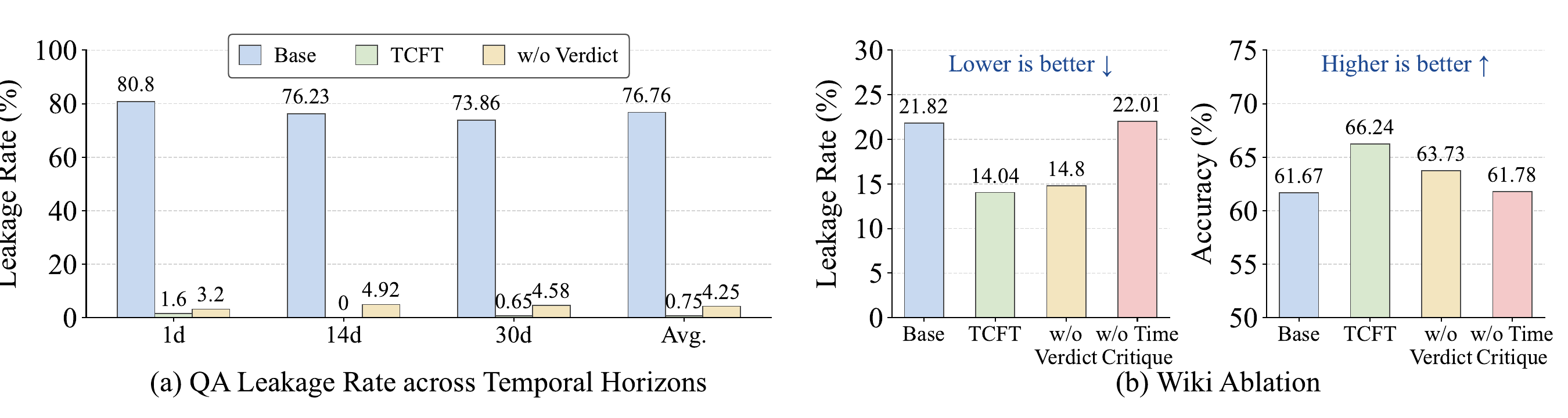}
    \caption{Ablation results of TCFT on QA and Wiki with Qwen2.5-7B-Instruct.}
    \label{fig:abla}
\end{figure}
\subsection{Ablation Study of TCFT (RQ3)}
We conduct ablation experiments to analyze the effects of the verdict signal $v$ and the rationale signal $r$ in TCFT. For QA, we remove the verdict signal, yielding the \textit{w/o Verdict} variant. For Wiki, we ablate both the verdict and rationale components. Since Wiki requires multi-fact generation, we replace the time-aware rationale with a generic hallucination critique, which checks factuality but does not verify whether each fact is supported before the cutoff. This \textit{w/o Time Critique} variant isolates the contribution of cutoff-aware reasoning. Figure~\ref{fig:abla} presents the ablation results. We can observe that:
\begin{itemize}[leftmargin=*]
\vspace{-5pt}
    \item \textbf{Temporal-validity rationales are essential for reliable ex-ante generation.}
    Replacing the time-aware critique with a generic factuality critique increases Wiki leakage from 14.04\% to 22.01\% and reduces accuracy from 66.24\% to 61.78\%, nearly reverting to the base model. This shows that factual correctness alone is insufficient for ex-ante generation, since true facts may still be invalid under the cutoff constraint. Cutoff-aware critique helps the model distinguish factual validity from temporal validity, enabling it to retain pre-cutoff facts while filtering out post-cutoff knowledge.
    \vspace{-3pt}
    \item \textbf{The verdict signal provides useful supervision for identifying cutoff violations.}
    Removing the verdict signal increases average QA leakage from 0.75\% to 4.25\%, with consistent degradation across all cutoff gaps. On Wiki, leakage also rises from 14.04\% to 14.80\%, while accuracy drops from 66.24\% to 63.73\%. These results show that the verdict signal helps models identify cutoff violations. However, the relatively mild degradation suggests that binary judgment alone provides limited supervision; robust ex-ante generation requires learning the underlying temporal verification rationale, \ie the ability to verify whether a fact is temporally valid under the cutoff constraint.
    \vspace{-5pt}
\end{itemize}

%% file: table/qa_wiki_results.tex
\begin{tabular}{
ccccc
>{\hspace{1.4em}}c c
}
\toprule
\multirow{2}{*}{\textbf{Models / Methods}}
& \multicolumn{4}{c}{\textbf{QA}}
& \multicolumn{2}{c}{\textbf{Wiki}} \\
\cmidrule(lr){2-5}
\cmidrule(l){6-7}
& \textbf{1d \textcolor{blue}{$\downarrow$}}
& \textbf{14d \textcolor{blue}{$\downarrow$}}
& \textbf{30d \textcolor{blue}{$\downarrow$}}
& \textbf{Avg. \textcolor{blue}{$\downarrow$}}
& \textbf{Leakage Rate \textcolor{blue}{$\downarrow$}}
& \textbf{Accuracy \textcolor{red}{$\uparrow$}} \\
\midrule

GPT-5.4
& 49.60 & 13.11 & 7.19 & 22.25
& 13.28 & 81.37 \\

Claude-Opus-4.6
& 48.00 & 11.48 & 5.23 & 20.50
& 16.86 & 81.30 \\

Gemini-3.1-Pro-Preview
& 44.00 & 41.80 & 40.52 & 42.00
& 15.75 & 79.82 \\

\midrule
\rowcolor{gray!15}
\multicolumn{7}{c}{\textbf{Qwen-2.5-7B-Instruct}} \\

Zero-shot
& 80.80 & 76.23 & 73.86 & 76.75
& 21.82 & 61.67 \\

Few-shot
& 20.80 & 26.23 & 18.95 & 21.75
& 26.76 & 59.96 \\

Chain-of-thought
& 59.20 & 58.20 & 51.63 & 56.00
& 22.77 & 62.35 \\

Self-verification
& 30.40 & 33.61 & 25.49 & 29.50
& 22.01 & 62.81 \\

SFT
& 35.20 & 41.80 & 30.72 & 35.50
& 20.79 & 62.18 \\

\rowcolor{ourscyan}
\textbf{Ours}
& \textbf{1.60} & \textbf{0.00}
& \textbf{0.65} & \textbf{0.75}
& \textbf{14.04} & \textbf{66.24} \\

\midrule
\rowcolor{gray!15}
\multicolumn{7}{c}{\textbf{Qwen-2.5-14B-Instruct}} \\

Zero-shot
& 81.44 & 65.66 & 62.28 & 69.80
& 23.34 & 69.49 \\

Few-shot
& 23.35 & 20.48 & 13.77 & 19.20
& 30.93 & 66.68 \\

Chain-of-thought
& 74.25 & 56.02 & 56.29 & 62.20
& 23.15 & 67.86 \\

Self-verification
& 24.55 & 11.45 & 9.58 & 15.20
& 20.49 & 70.73 \\

SFT
& 20.80 & 25.41 & 17.00 & 20.75
& 16.32 & 71.01 \\

\rowcolor{ourscyan}
\textbf{Ours}
& \textbf{5.99} & \textbf{6.02}
& \textbf{5.39} & \textbf{5.80}
& \textbf{11.76} & \textbf{73.81} \\

\bottomrule
\end{tabular}

%% file: chapter/5_conclusion.tex
\section{Conclusion}
In this work, we study ex-ante reasoning as a temporal verification problem, where models must judge whether a response is admissible under a given cutoff. We show that prompt-level interventions provide brittle control, with leakage remaining sensitive to formulation and placement. To address this, we propose TCFT, a Temporal Critique Fine-Tuning framework that trains models to identify post-cutoff leakage, explain boundary violations, and judge temporal admissibility. Experiments on QA and Wiki show that TCFT reduces leakage while preserving factual utility. Ablation and cross-template studies further show that TCFT learns transferable cutoff-aware verification beyond prompt-specific patterns. Overall, reliable ex-ante reasoning requires verification rather than stronger prompting or answer imitation.
\section{Limitations}
While TCFT provides an effective framework for cutoff-aware verification, limitations remain. Our evaluation focuses on QA and Wiki-style generation, leaving broader time-sensitive scenarios such as finance, law, and scientific trend analysis unexplored. We use Qwen2.5-7B-Instruct and Qwen2.5-14B-Instruct, as many smaller open-weight models weakly memorize post-cutoff facts, making leakage evaluation and fine-tuning less informative. Extending TCFT to stronger backbones and reasoning-specialized models remains future work. TCFT also relies on teacher-generated temporal critiques, whose quality may vary across teacher models, critique styles, and evidence sources; developing more robust critique-generation strategies is another important direction.

\section*{Acknowledgements}
This research is supported by the National Natural Science Foundation of China (62625608, 62572449, 62302321) and the Fundamental Research Funds for the Central Universities (WK2100250065).
The work of Yancheng Yuan was supported by the NSFC Young Scientists Fund (Project No. 12501440).
This research is also supported by the advanced computing resources provided by the Supercomputing Center of the USTC.

%% file: chapter/7_appendix.tex
\appendix
\section{Related Work}
\subsection{Temporal Reasoning}
Evaluating LLMs on time-related dimensions has historically centered on temporal reasoning and time-sensitive question answering~\cite{temporal1,temporal2}.
Benchmarks such as TRAM~\cite{TRAM}, TimeBench~\cite{TimeBench}, and Test of Time (ToT)~\cite{TestOfTime} rigorously assess models on event ordering, durations, and temporal arithmetic~\cite{ExAnte,Dated_Data}.
Similarly, time-sensitive QA datasets evaluate how models adapt to evolving knowledge~\cite{StreamingQA}.
However, these existing benchmarks generally assume full access to future context or focus on static facts, largely overlooking the specific requirement of cutoff-aware temporal restriction~\cite{time1,time2}.

To bridge this gap, the ExAnte benchmark~\cite{ExAnte} was introduced to systematically evaluate LLMs' ability to reason while strictly adhering to temporal cutoffs.
Experimental results across multiple tasks reveal that existing LLMs frequently violate these temporal constraints, with no single prompting strategy—including Zero-shot, Chain-of-Thought (CoT)~\cite{cot}, and Self-Verification~\cite{self-verification}—consistently eliminating temporal leakage~\cite{ExAnte}.
Beyond general failure, empirical observations indicate specific vulnerabilities: (1) \textbf{Cutoff Explicitness Effect}: Models exhibit a significant sensitivity to whether a cutoff is phrased through implicit or explicit templates, suggesting a reliance on surface-level linguistic cues rather than robust time-conditioned validity reasoning.
(2) \textbf{Constraint Placement Sensitivity}: A distinct ``prefix advantage'' is observed, where temporal constraints at the beginning of a prompt yield more effective control than those at the end—a phenomenon that contradicts the intuition that proximity to the output should enhance constraint adherence.
These insights suggest that failures in ex-ante reasoning are fundamentally temporal verification failures, indicating a lack of genuine temporal verification capability in current models.
Since surface-level prompting proves insufficient to resolve these internal reasoning gaps, research has begun to explore more intrusive forms of knowledge control~\cite{InContextUnlearning,unlearningsurvey,llmeraser}.

\subsection{LLM Unlearning}
As prompting strategies fail to provide a robust solution for temporal adherence, machine unlearning has emerged as a potential paradigm to suppress or remove specific knowledge within model weights~\cite{unlearningsurvey,SaferUnlearning,llmeraser}.
Subsequent research has proposed various frameworks, such as gradient-ascent fine-tuning~\cite{UnrollingSGD,task}, preference-based unlearning~\cite{NPO}, and in-context unlearning~\cite{InContextUnlearning}, to selectively erase the influence of specific data points \cite{mllmeraser,ding2026cuse}.

However, ex-ante reasoning presents a challenge fundamentally distinct from traditional unlearning.
While unlearning focuses on the permanent removal of data for privacy or safety, ex-ante inference requires a dynamic, context-dependent suppression of knowledge based on an arbitrary temporal cutoff~\cite{ExAnte}.
A critical obstacle is the ``memorization-leakage'' paradox: evidence suggests that stronger recall and memorization of post-cutoff events are often positively correlated with higher leakage rates, as models struggle to decouple internalized memories from temporal constraints~\cite{ExAnte,Lookaheadbias}.
Consequently, ex-ante reasoning demands temporary knowledge suppression and active admissibility judgment—capabilities that traditional unlearning methods, which target permanent erasure, are not designed to achieve~\cite{ExAnte}.

\subsection{Critique Learning}
To overcome the limitations of standard prompting and Supervised Fine-Tuning (SFT) \cite{instrcut,Self-Instruct,sft,ft}, recent research has turned toward self-correction and critique learning~\cite{Self-Refine,Reflexion,gou2023critic}.
While self-correction aims to refine outputs, studies show that LLMs typically struggle to identify complex flaws without specialized training~\cite{xu2025double}.
Critique Fine-Tuning (CFT) was proposed to shift the training paradigm from simple imitation to active verification, training models to critique noisy responses and identify errors~\cite{CritiqueFineTuning,valmeekam2023can}.

In the context of ex-ante reasoning, standard SFT is inadequate because its objective is static imitation; yet, in ex-ante scenarios, the validity of a response is strictly conditioned on the specified temporal cutoff.
This necessitates an extension of the critique paradigm to the temporal domain.
By training models to evaluate the temporal admissibility of responses and identify specific leakage violations, it is possible to move beyond superficial prompt adherence.
Such an approach aims to instill an intrinsic cutoff-aware verification capability, enabling models to judge whether a response remains valid under shifting temporal constraints~\cite{Verify_step,judgement,wang2024math,yuan2024free,ankner2024critique}.


\section{Validity Evaluation Metric}
\label{app:validity_metric}
Another core dimension of evaluating ex-ante reasoning is guarding against degenerate low-leakage strategies~\cite{ExAnte}.
For instance, a naive model could trivially achieve zero leakage by refusing to answer (\eg \textit{``I cannot provide the related information''}) or by generating vacuous, low-quality text, which while technically adhering to the cutoff, offers no functional utility.

To distinguish genuine reasoning from mere abstention, a quality guardrail $Q(\hat{y}, y^*)$ is introduced, where $Q$ is a task-specific similarity or factual precision function evaluated against the ideal pre-cutoff response~\cite{ExAnte}.
The instantiation of $Q$ is task-dependent to reflect diverse output structures~\cite{ExAnte}.
For numerical single-event tasks, the quality function can be instantiated with task-specific regression metrics such as negative mean absolute error~\cite{ExAnte}. 
For QA, we do not evaluate validity separately, since the task asks for the exact date of an unexpected future event, and providing the correct date is itself temporal leakage. 
Thus, there is no meaningful pre-cutoff target answer for quality evaluation; rewarding exact correctness would reward leakage, while penalizing uncertainty or abstention would be inappropriate. 
We therefore evaluate QA using leakage rate only.

In the context of multi-event generation, $Q$ is operationalized as factual precision, representing the proportion of generated atomic claims that are verifiable within the $R_{\text{prev}}$ knowledge space~\cite{ExAnte, Min2023FactScore}. Formally, a response is \textit{valid} only if it satisfies both the temporal boundary and a minimum utility threshold $\gamma$~\cite{ExAnte}:
\begin{equation}
    \text{Valid}(x) = \mathbf{1}[L(x) = 0] \land \mathbf{1}[Q(\hat{y}, y^*) \geq \gamma].
\end{equation}

\section{Discussion}
\subsection{Relation to Supervised Fine-Tuning}
\label{sec:method_vs_sft}
Formally, standard SFT \cite{Self-Instruct,sft,ft,instrcut} optimizes the following objective:
\[
\min_\theta \mathbb{E}_{(x,t_c,y^*)} -\log p(y^* \mid x,t_c;\theta),
\]
where $y^*$ denotes the oracle response under cutoff $t_c$.
This objective treats $(x,t_c)$ as the input context and $y^*$ as the target output, thereby learning a direct mapping from prompt to answer.
However, because $y^*$ is constructed to be temporally admissible by design, SFT only exposes the model to positive answers and provides no supervision on what future leakage looks like or why it violates the cutoff.
As a result, it remains an answer-generation objective rather than a temporal verification objective.

TCFT directly addresses this limitation.
Instead of training only on admissible target answers, TCFT pairs diverse candidate responses, including both admissible and leaked ones, with explicit temporal judgments.
The model therefore learns to evaluate a response against the cutoff, distinguishing temporally valid content from post-cutoff leakage, rather than merely memorizing a mapping from context to answer.

\subsection{Relation to Conventional Critique Fine-Tuning}
\label{sec:method_vs_cft}
Template CFT~\cite{CritiqueFineTuning} has shown promise in teaching models to identify errors in problem-solving chains and self-correct.
Its key contribution is to shift learning from output imitation to error identification.
TCFT extends this inversion to the temporal setting, where the object of critique is not only whether a response is factually correct, but whether it is admissible under a given cutoff.

In standard CFT, the critique of a reasoning error is largely context-independent: whether a solution step is flawed depends on the problem and the step itself.
In temporal critique, however, admissibility is cutoff-dependent.
For example, the statement ``Elon Musk acquired Twitter'' is factually correct, but temporally inadmissible if the cutoff precedes 2022.
Likewise, the same response $y$ may be valid when $t_c=2023$ but invalid when $t_c=2020$.
Thus, the critique function must be conditioned on the cutoff:
\begin{equation}
	c^{(t_c)} = \mathcal{C}_\theta(x,t_c,y) \neq \mathcal{C}_\theta(x,y).
\end{equation}
This makes the learned critique capability inherently temporally parameterized.
Without conditioning on $t_c$, a CFT-style critique would assign the same judgment across different cutoffs, failing to capture the temporal validity relation central to ex-ante reasoning.

\section{Experimental Setups}
\label{app:experimental_setups}

\subsection{Datasets}
\label{app:datasets}
We build our experiments on the ExAnte benchmark~\citep{ExAnte}, focusing on two representative ex-ante inference tasks: single-event QA and Wikipedia-style generation. These two tasks cover different levels of response complexity: QA requires the model to decide whether it can answer a future-event question under a given cutoff, while Wiki requires the model to generate multiple factual claims that must all remain valid before the cutoff. To improve generalization, we augment both datasets along several dimensions. Specifically, we vary whether constraints are expressed implicitly or explicitly, diversify constraint positions in the prompt, and introduce different cutoff-date settings. We further mix multiple prompt templates during data construction, so the model is exposed to semantically similar tasks with varied surface forms rather than overfitting to a single prompt pattern. The QA dataset contains 9,072 examples in total, split into 8:1:1 for train/validation/test. The Wiki dataset contains 5,268 examples in total, split into 8:1:1 for train/validation/test.

\paragraph{QA.}
The QA task evaluates single-event temporal leakage. Each example contains an event query, the true event date, and a cutoff date preceding the event. Following our main experiments, we evaluate three cutoff gaps: 1 day, 14 days, and 30 days before the target event date. A response is counted as leaked if the model reveals the correct post-cutoff event information under the specified cutoff. For TCFT training, each QA example is augmented along three dimensions: cutoff gap, temporal formulation, and constraint placement. Specifically, we use three cutoff gaps, two formulations (implicit historical framing and explicit knowledge restriction), and two placements (prefix and suffix), yielding $3 \times 2 \times 2 = 12$ template variants per original QA example.

\paragraph{Wiki.}
The Wiki task evaluates multi-event generation under temporal cutoffs. Given a topic and a cutoff year, the model is asked to generate factual claims using only information available before the cutoff. Since Wiki-style prompts often involve long-span entities or events, we keep the cutoff fixed for each example and augment the prompt along two dimensions: temporal formulation and constraint placement. This yields $2 \times 2 = 4$ template variants per original Wiki example. During evaluation, generated responses are decomposed into atomic claims, and each claim is checked for whether it is supported before the cutoff or leaks post-cutoff information.

\subsection{Evaluation Metrics}
\label{app:evaluation_metrics}

We use leakage rate as the primary metric for both QA and Wiki. Lower leakage indicates stronger cutoff adherence.

\paragraph{QA leakage.}
For QA, leakage is measured under 1-day, 14-day, and 30-day cutoff gaps. A response is considered leaked if it provides the correct post-cutoff event information despite being instructed to answer under the cutoff. We report leakage rates for each cutoff gap and their average.

\paragraph{Wiki leakage and accuracy.}
For Wiki, we report leakage rate and claim accuracy. Leakage rate measures the proportion of generated claims that rely on post-cutoff information. Claim accuracy measures the proportion of generated claims that are factually supported by pre-cutoff information. This metric is included to prevent degenerate low-leakage strategies, such as refusing to answer or producing overly vague content. Detailed definitions of leakage rate and claim accuracy are provided in Table~\ref{tab:claim_eval}.

\begin{table}[t]
\centering
\small
\caption{Claim-level evaluation for Wiki generation.}
\label{tab:claim_eval}
\begin{tabular}{c c c c}
\toprule
\textbf{Pre-cutoff support} & \textbf{Post-cutoff support} & \textbf{Temporal leakage} & \textbf{Claim accuracy} \\
\midrule
$f_i \in R_{\mathrm{pre}}$ & $f_i \notin R_{\mathrm{post}}$ & No & Yes \\
$f_i \notin R_{\mathrm{pre}}$ & $f_i \in R_{\mathrm{post}}$ & Yes & No \\
$f_i \notin R_{\mathrm{pre}}$ & $f_i \notin R_{\mathrm{post}}$ & No & No \\
$f_i  \in R_{\text{prev}}$ & $f_i  \in R_{\text{post}}$ & No  & Yes \\
\bottomrule
\end{tabular}
\end{table}

\subsection{Baselines}
\label{app:baselines}

We compare TCFT with proprietary LLMs, prompt-level interventions, and a parameter-level fine-tuning baseline.

\paragraph{Proprietary LLMs.}
We evaluate GPT-5.4~\cite{OpenAI2026GPT54}, Claude-Opus-4.6~\cite{Anthropic2026ClaudeOpus46}, and Gemini-3.1-Pro-Preview~\cite{Google2026Gemini31ProPreview} under the same evaluation protocol. These models serve as strong proprietary references for assessing whether general model capability alone leads to reliable cutoff adherence.

\paragraph{Zero-shot.}
The model is directly prompted with the task query and the cutoff constraint, without additional demonstrations or verification steps.

\paragraph{Few-shot~\cite{brown2020language}.}
The model is provided with demonstration examples that illustrate how to answer under temporal cutoffs. The demonstrations are designed to encourage the model to avoid post-cutoff information while preserving useful pre-cutoff content.

\paragraph{Chain-of-thought~\cite{cot}.}
The model is prompted to reason step by step before answering. This baseline tests whether eliciting intermediate reasoning improves temporal cutoff adherence.

\paragraph{Self-verification~\cite{self-verification}}.
The model first generates an initial response and then performs an additional verification step to check whether the response contains post-cutoff information. If leakage is detected, the model is instructed to revise the response.

\paragraph{SFT.}
We train a supervised fine-tuning baseline on temporally admissible target answers generated by GPT-5.4~\cite{OpenAI2026GPT54} for each query under the corresponding cutoff. This baseline uses the same backbone models and augmented prompt formats as TCFT, but optimizes final-answer imitation rather than temporal critique~\cite{sft,ft,instrcut}. It is included to test whether answer-level supervision alone is sufficient for learning ex-ante reasoning.

\subsection{Prompt Templates}
\label{app:prompt_templates}

We use two types of temporal formulations and two constraint placements. We present several representative template examples here, with the complete set shown in Table~\ref{tab:qa_templates}. In practice, training incorporates diverse temporal templates to enhance generalization across different cutoff formulations.

\paragraph{Implicit temporal formulation.}
The cutoff is expressed through a historical framing:
\begin{quote}
Assume today is \texttt{<cutoff date>}. \texttt{<query>}
\end{quote}

\paragraph{Explicit temporal formulation.}
The cutoff is expressed as a direct knowledge restriction:
\begin{quote}
Use only information available before \texttt{<cutoff date>}. \texttt{<query>}
\end{quote}

\paragraph{Prefix placement.}
The temporal constraint is placed before the query:
\begin{quote}
Use only information available before \texttt{<cutoff date>}. \texttt{<query>}
\end{quote}

\paragraph{Suffix placement.}
The temporal constraint is placed after the query:
\begin{quote}
\texttt{<query>} Use only information available before \texttt{<cutoff date>}.
\end{quote}

\begin{table}[t]
\centering
\small
\caption{Representative explicit and implicit temporal templates for the QA task, where \{Q\} denotes the input query.}
\label{tab:qa_templates}
\renewcommand{\arraystretch}{1.18}
\setlength{\tabcolsep}{5pt}
\arrayrulecolor{gray!35}
\begin{tabularx}{\textwidth}{
>{\centering\arraybackslash}p{0.06\textwidth}
>{\raggedright\arraybackslash}X
>{\raggedright\arraybackslash}X
}
\toprule
\textbf{ID} & \textbf{Explicit template} & \textbf{Implicit template} \\
\midrule

\rowcolor{gray!6}
01 &
\textbf{Constrain your response to pre-$t_c$ information only. Any information beyond $t_c$ must be excluded.} \{Q\} 
\newline
\newline
You should provide answer follow the format of YYYY-MM-DD. If you don't know, please answer I don't know.
&
\textbf{Assume today is $t_c$.} \{Q\} 
\newline
\newline
You should provide answer follow the format of YYYY-MM-DD. If you don't know, please answer I don't know. \\
\midrule

02 &
\textbf{Base your response entirely on information available by $t_c$, without using any later information.} \{Q\} 
\newline
\newline
You should provide answer follow the format of YYYY-MM-DD. If you don't know, please answer I don't know.
&
\textbf{Suppose you are answering this question on $t_c$.} \{Q\} 
\newline
\newline
You should provide answer follow the format of YYYY-MM-DD. If you don't know, please answer I don't know. \\
\midrule

\rowcolor{gray!6}
03 &
\textbf{Your response must rely only on information available on or before $t_c$. Any later information must not be used.} \{Q\} 
\newline
\newline
You should provide answer follow the format of YYYY-MM-DD. If you don't know, please answer I don't know.
&
\textbf{Write your answer as it would be given on $t_c$.} \{Q\} 
\newline
\newline
You should provide answer follow the format of YYYY-MM-DD. If you don't know, please answer I don't know. \\
\midrule

04 &
\textbf{Restrict your answer to information available on or before $t_c$ only. Do not include any facts that were not available by $t_c$.} \{Q\} 
\newline
\newline
You should provide answer follow the format of YYYY-MM-DD. If you don't know, please answer I don't know.
&
\textbf{Provide your answer as of $t_c$. }\{Q\} 
\newline
\newline
You should provide answer follow the format of YYYY-MM-DD. If you don't know, please answer I don't know. \\
\midrule

\rowcolor{gray!6}
05 &
\textbf{Your answer must be based exclusively on information available by $t_c$. Do not include any facts from after $t_c$.} \{Q\} 
\newline
\newline
You should provide answer follow the format of YYYY-MM-DD. If you don't know, please answer I don't know.
&
\textbf{Frame your answer as of $t_c$.} \{Q\} 
\newline
\newline
You should provide answer follow the format of YYYY-MM-DD. If you don't know, please answer I don't know. \\
\bottomrule
\end{tabularx}
\arrayrulecolor{black}
\end{table}
\subsection{Implementation Details}
\label{app:implementation_details}

We use Qwen2.5-7B-Instruct and Qwen2.5-14B-Instruct \cite{qwen-2.5} as the open-weight backbones. We exclude smaller models because their limited memorization of post-cutoff facts makes leakage evaluation less informative. All models are evaluated on the same QA and Wiki test sets.

\paragraph{Temporal critique generation.}
For TCFT, we construct critique supervision using a teacher model (\ie GPT-5.4 \cite{OpenAI2026GPT54}). Given a query, cutoff, and candidate response, the teacher generates a temporal-validity rationale and a binary verdict. The rationale explains which parts of the response are valid or invalid under the cutoff, while the verdict summarizes whether the response contains post-cutoff leakage.

\paragraph{Training.}
TCFT is trained using standard supervised fine-tuning over critique-conclusion sequences. The input consists of the query, cutoff, and candidate response, while the target output consists of the temporal rationale and verdict. We optimize the model with the standard next-token prediction objective. For fair comparison, SFT uses the same backbone and prompt augmentation but is trained to imitate final target answers rather than critiques.

\paragraph{Inference.}
At inference time, TCFT directly generates the final response from the query and cutoff. It does not require explicit critique generation or answer repair during evaluation. The learned temporal verification capability is expected to guide generation away from post-cutoff information.

\paragraph{Hyperparameters.}
\label{appendix:Hyperparameters}
Unless otherwise stated, we use the same training configuration for SFT and
TCFT. The learning rate is set to \texttt{1e-5}, warmup ratio to \texttt{0.03}. The effective batch size
is \texttt{64}, implemented with a per-device batch size of \texttt{16},
gradient accumulation steps of \texttt{2}, and \texttt{2} GPUs. We train
for \texttt{3} epochs with a maximum sequence length of \texttt{1024}.
Training uses \texttt{bfloat16} precision and gradient checkpointing. All
experiments use a fixed random seed of \texttt{42}. For LoRA, we use rank
\texttt{16}, LoRA alpha \texttt{32}, and LoRA dropout \texttt{0.05}, and
apply adapters to \texttt{q\_proj}, \texttt{k\_proj}, \texttt{v\_proj},
\texttt{o\_proj}, \texttt{gate\_proj}, \texttt{up\_proj}, and
\texttt{down\_proj}. Experiments are run on a server with NVIDIA A100 GPUs (\texttt{80} GB memory per GPU).

\section{Prompt Templates}
\subsection{Temporal Critique Prompt for QA}

\begin{systempromptbox}[title={System Prompt}]
\small
You are an expert critic for ex-ante temporal reasoning.

You will receive an event, its true date, a cutoff date, the original prompt, and a model answer. Your job is to critique the answer under the cutoff constraint.

\textbf{Rules:}
\begin{itemize}[leftmargin=*]
    \item If the true event date is later than the cutoff date, then the event had not happened yet at the cutoff. Any answer that gives a specific future date or otherwise relies on post-cutoff knowledge is wrong.
    \item If the true event date is on or before the cutoff date, then the date was already knowable at the cutoff. In this case, the answer is right only if it gives the correct date in the required format. If it is wrong, explain the factual mistake. In this case, it is allowed to mention the correct date in the critique.
    \item Focus first on temporal validity, then on factual correctness, then on format.
    \item Keep the critique concise, natural, and useful for training.
    \item End with exactly one of: ``Conclusion: right'' or ``Conclusion: wrong''.
\end{itemize}
\end{systempromptbox}

\begin{userpromptbox}[title={User Prompt}]
\small
Event: \{event\}

True event date: \{true\_date\}

Cutoff date: \{ask\_date\}

Original prompt:
\{leakage\_prompt\}

Model answer:
\{leakage\_answer\_raw\}

Write a concise critique of the model answer under the cutoff date.
\end{userpromptbox}

\subsection{Temporal Critique Prompt for Wiki}
\label{app:wiki_critique_prompt}

\begin{systempromptbox}[title={System Prompt}]
\small
You are an expert critic for ex-ante temporal reasoning on Wikipedia-style factual claims.

You will receive a topic, a cutoff date, the original prompt, and a model answer containing several atomic claims. Your job is to critique whether the claims are valid under the cutoff constraint.

\textbf{Rules:}
\begin{itemize}[leftmargin=*]
    \item Evaluate each claim according to what could reasonably be known on or before the cutoff date.
    \item A claim is wrong if it relies on information from after the cutoff date.
    \item For future events, distinguish scheduled or planned information from completed-event information:
    \begin{itemize}[leftmargin=*]
        \item It is acceptable to say that an event was scheduled, planned, announced, selected, or expected before the cutoff if that was knowable then.
        \item It is not acceptable to state a future event as already having happened, completed, concluded, won, opened, released, died, etc., if that happened after the cutoff.
    \end{itemize}
    \item If any claim is temporally invalid, the overall answer is wrong.
    \item If all claims are temporally valid and factually plausible under the cutoff, the overall answer is right.
    \item Keep the critique concise, natural, and useful for training.
    \item End with exactly one of: ``Conclusion: right'' or ``Conclusion: wrong''.
\end{itemize}
\end{systempromptbox}

\begin{userpromptbox}[title={User Prompt}]
\small
Topic: \{title\}

Cutoff date: \{cutoff\_date\}

Original prompt:
\{prompt\}

Model answer:
\{raw\_response\}

Claims:
\{claims\}

Write a concise critique of the model answer under the cutoff date.
\end{userpromptbox}

\subsection{Example of Wiki Temporal Critique Data}
\label{app:wiki_critique_example}

\begin{sampleinputbox}[title={Input}]
\small
\textbf{Original prompt:}

Regarding your knowledge about ``2020\_Summer\_Olympics'', generate 5 atomic facts. Each fact should be a short, clear statement representing a distinct piece of information. Do not guess or fill in missing details. Format each fact as a numbered list (1--5). Frame your answer as of January 1st, 2020.

\medskip
\textbf{Model answer:}

1. The 2020 Summer Olympics were scheduled to take place in Tokyo, Japan.

2. The opening ceremony of the 2020 Summer Olympics was planned for July 24, 2020.

3. The 2020 Summer Olympics were originally set to include 33 sports across 339 events.

4. The Olympic Village for the 2020 Summer Olympics was located in Musashino Forest Park.

5. The torch relay for the 2020 Summer Olympics began on March 26, 2020.
\end{sampleinputbox}

\begin{samplelabelbox}[title={Critique}]
\small
1--3 are fine as of Jan. 1, 2020: Tokyo was the host city, the opening ceremony was scheduled for July 24, 2020, and the program was planned for 33 sports and 339 events.

\medskip
4 is wrong factually. Musashino Forest Park was a venue, not the Olympic Village. The Olympic Village was in the Harumi waterfront area of Tokyo.

\medskip
5 is not valid under the prompt's time framing. As of Jan. 1, 2020, one could say the torch relay was scheduled/planned to begin on March 26, 2020, but saying it ``began'' states a future event as already occurred.

\medskip
\textbf{Conclusion: wrong}
\end{samplelabelbox}

\subsection{Example of QA Temporal Critique Data}
\label{app:qa_critique_example}

\begin{sampleinputbox}[title={Input}]
\small
\textbf{Original prompt:}

Your response must rely only on information available on or before 2018-06-10. Any later information must not be used. Now please answer the following question. Just leave the best answer without explanation.

Question: On what date Twelve boys and their football coach are rescued from the flooded Tham Luang Nang Non cave in Thailand? You should provide answer follow the format of YYYY-MM-DD. If you don't know, please answer I don't know. 

\medskip
\textbf{Model answer:}

2018-07-10
\end{sampleinputbox}

\begin{samplelabelbox}[title={Critique}]
\small
The answer uses post-cutoff information: the rescue had not happened yet by 2018-06-10, so the exact rescue date was not knowable at the cutoff. It should have abstained and answered ``I don't know'' instead of giving a future date.

\medskip
\textbf{Conclusion: wrong}
\end{samplelabelbox}

\subsection{Wiki Critique Prompt w/o temporal critique}
\label{app:wiki_factual_critique_prompt}

\begin{systempromptbox}[title={System Prompt}]
\small
You are an expert critic for Wikipedia-style factual claims.

You will receive a title, an original prompt, and a model answer containing several atomic claims. Your job is to critique whether the model answer is factually correct.

\textbf{Rules:}
\begin{itemize}[leftmargin=*]
    \item Evaluate each claim according to general factual correctness.
    \item A claim is wrong if it is factually incorrect, unsupported, or overly specific in an unsupported way.
    \item Prefer concise, natural critiques useful for training.
    \item End with exactly one of: ``Conclusion: right'' or ``Conclusion: wrong''.
\end{itemize}
\end{systempromptbox}

\begin{userpromptbox}[title={User Prompt}]
\small
Title: \{title\}

Cutoff year: \{cutoff\_year\}

Original prompt:

\{prompt\}

Model answer:

\{model\_answer\}

Write a concise factual critique of the model answer.
\end{userpromptbox}

\subsection{Wiki Critique Prompt w/o verdict}
\label{app:wiki_critique_prompt}

\begin{systempromptbox}[title={System Prompt}]
\small
You are an expert critic for ex-ante temporal reasoning on Wikipedia-style factual claims.

You will receive a topic, a cutoff date, the original prompt, and a model answer containing several atomic claims. Your job is to critique whether the claims are valid under the cutoff constraint.

\textbf{Rules:}
\begin{itemize}[leftmargin=*]
    \item Evaluate each claim according to what could reasonably be known on or before the cutoff date.
    \item A claim is wrong if it relies on information from after the cutoff date.
    \item For future events, distinguish scheduled or planned information from completed-event information:
    \begin{itemize}[leftmargin=*]
        \item It is acceptable to say that an event was scheduled, planned, announced, selected, or expected before the cutoff if that was knowable then.
        \item It is not acceptable to state a future event as already having happened, completed, concluded, won, opened, released, died, etc., if that happened after the cutoff.
    \end{itemize}
    \item If any claim is temporally invalid, the overall answer is wrong.
    \item If all claims are temporally valid and factually plausible under the cutoff, the overall answer is right.
    \item Keep the critique concise, natural, and useful for training.
\end{itemize}
\end{systempromptbox}

\section{Case Study}
\begin{userpromptbox}[title={User Prompt}]
\small
Topic: \{title\}

Cutoff date: \{cutoff\_date\}

Original prompt:
\{prompt\}

Model answer:
\{raw\_response\}

Claims:
\{claims\}

Write a concise critique of the model answer under the cutoff date.
\end{userpromptbox}
\begin{casepromptbox}[title={Prompt}]
\footnotesize
Assume today is January 1st, 2011. Regarding your knowledge about ``Emma\_Watson'', generate 5 atomic facts. Each fact should be a short, clear statement representing a distinct piece of information. Do not guess or fill in missing details. Format each fact as a numbered list (1--5).
\end{casepromptbox}

\begin{baselinebox}[title={Zero-shot}]
\footnotesize
\begin{enumerate}[leftmargin=1.3em,itemsep=1pt,topsep=1pt]
    \item Emma Watson was born on April 15, 1990.
    \item Emma Watson played the character Hermione Granger in the Harry Potter film series.
    \item Emma Watson is an English actress and \textcolor{darkred}{UN Women Goodwill Ambassador \emph{(appointed in Jul.~2014)}}.
    \item Emma Watson \textcolor{darkred}{graduated from Brown University in 2014 \emph{(after the cutoff)}}.
    \item Emma Watson has blonde hair and blue eyes.
\end{enumerate}
\end{baselinebox}

\begin{baselinebox}[title={Few-shot}]
\footnotesize
\begin{enumerate}[leftmargin=1.3em,itemsep=1pt,topsep=1pt]
    \item Emma Watson was born on April 15, 1990.
    \item She played the role of Hermione Granger in the Harry Potter film series.
    \item Emma Watson was \textcolor{darkred}{appointed as a UN Women Goodwill Ambassador in 2014 \emph{(after the cutoff)}}.
    \item She \textcolor{darkred}{graduated from Brown University in 2017 with a degree in English literature and French studies \emph{(post-cutoff; actual graduation was in 2014)}}.
    \item As of January 1st, 2011, she was known for her roles in the Harry Potter films and had not yet become a UN Women Goodwill Ambassador.
\end{enumerate}
\end{baselinebox}

\begin{baselinebox}[title={Chain-of-thought}]
\footnotesize
\begin{enumerate}[leftmargin=1.3em,itemsep=1pt,topsep=1pt]
    \item Emma Watson was born on April 15, 1990.
    \item Emma Watson played the character Hermione Granger in the Harry Potter film series.
    \item Emma Watson is an English actress and \textcolor{darkred}{UN Women Goodwill Ambassador \emph{(appointed in Jul.~2014)}}.
    \item Emma Watson \textcolor{darkred}{graduated from Brown University in 2014 \emph{(after the cutoff)}}.
    \item Emma Watson has blonde hair and blue eyes.
\end{enumerate}
\end{baselinebox}

\begin{baselinebox}[title={Self-verification}]
\footnotesize
\begin{enumerate}[leftmargin=1.3em,itemsep=1pt,topsep=1pt]
    \item Emma Watson was born on April 15, 1990.
    \item On January 1, 2011, Emma Watson was starring in the film ``The Perks of Being a Wallflower.''
    \item As of January 1, 2011, Emma Watson had completed her role as Hermione Granger in the Harry Potter film series.
    \item Emma Watson was on an \textcolor{darkred}{academic leave from Brown University \emph{(announced after the cutoff)}} as of January 1, 2011.
    \item By January 1, 2011, Emma Watson had been involved in various \textcolor{darkred}{UN Women campaigns promoting gender equality \emph{(UN Women involvement began after the cutoff)}}.
\end{enumerate}
\end{baselinebox}

\begin{ourscasebox}[title={Ours (TCFT)}]
\footnotesize
\begin{enumerate}[leftmargin=1.3em,itemsep=1pt,topsep=1pt]
    \item Emma Watson was born on April 15, 1990.
    \item She was known for her role as Hermione Granger in the Harry Potter film series.
    \item On January 1, 2011, she was 20 years old.
    \item She had appeared in the film ``Harry Potter and the Deathly Hallows -- Part 1'' before this date.
    \item She was a British actress.
\end{enumerate}
\end{ourscasebox}

%% file: checklist.tex
\section*{NeurIPS Paper Checklist}

\begin{enumerate}

\item {\bf Claims}
    \item[] Question: Do the main claims made in the abstract and introduction accurately reflect the paper's contributions and scope?
    \item[] Answer: \answerYes{}. 
    \item[] Justification: The abstract and introduction state the main problem of temporal leakage in ex-ante reasoning, analyze the brittleness of prompt-level control, and introduce TCFT as a temporal critique fine-tuning framework. The claimed improvements are supported by experiments on QA and Wiki tasks.

\item {\bf Limitations}
    \item[] Question: Does the paper discuss the limitations of the work performed by the authors?
    \item[] Answer: \answerYes{}. 
    \item[] Justification: The paper includes a limitations discussion, noting that the evaluation focuses on QA and Wiki-style generation, that many smaller open-weight models have limited memorization of post-cutoff facts, and that TCFT depends on teacher-generated temporal critiques.

\item {\bf Theory assumptions and proofs}
    \item[] Question: For each theoretical result, does the paper provide the full set of assumptions and a complete (and correct) proof?
    \item[] Answer: \answerNA{}. 
    \item[] Justification: The paper does not present formal theoretical results or theorems requiring proofs. The mathematical formulations are used to define ex-ante reasoning, leakage, and the TCFT objective.

    \item {\bf Experimental result reproducibility}
    \item[] Question: Does the paper fully disclose all the information needed to reproduce the main experimental results of the paper to the extent that it affects the main claims and/or conclusions of the paper (regardless of whether the code and data are provided or not)?
    \item[] Answer: \answerYes{}. 
    \item[] Justification: The paper describes the evaluated tasks, model backbones, baselines, metrics, data augmentation strategy, and training objective. Additional implementation details are provided in the appendix, and the code is released in an anonymized repository.

\item {\bf Open access to data and code}
    \item[] Question: Does the paper provide open access to the data and code, with sufficient instructions to faithfully reproduce the main experimental results, as described in supplemental material?
    \item[] Answer: \answerYes{}. 
    \item[] Justification: The paper uses the public ExAnte benchmark and provides the implementation in an anonymized repository. The repository includes instructions for data preprocessing, training, and evaluation. 

\item {\bf Experimental setting/details}
    \item[] Question: Does the paper specify all the training and test details (e.g., data splits, hyperparameters, how they were chosen, type of optimizer) necessary to understand the results?
    \item[] Answer: \answerYes{}. 
    \item[] Justification: The paper specifies the evaluated tasks, model backbones, cutoff settings, baselines, evaluation metrics, and data augmentation protocol. Additional training and implementation details are provided in the appendix.

\item {\bf Experiment statistical significance}
    \item[] Question: Does the paper report error bars suitably and correctly defined or other appropriate information about the statistical significance of the experiments?
    \item[] Answer: \answerNo{}. 
    \item[] Justification: The main experiments report leakage rates and accuracy scores, but do not include error bars, confidence intervals, or statistical significance tests. The evaluation is based on fixed benchmark splits and deterministic aggregate metrics.

\item {\bf Experiments compute resources}
    \item[] Question: For each experiment, does the paper provide sufficient information on the computer resources (type of compute workers, memory, time of execution) needed to reproduce the experiments?
    \item[] Answer: \answerYes{}. 
    \item[] Justification: The appendix reports the GPU type, training time, and total compute required for the main experiments.
    
\item {\bf Code of ethics}
    \item[] Question: Does the research conducted in the paper conform, in every respect, with the NeurIPS Code of Ethics \url{https://neurips.cc/public/EthicsGuidelines}?
    \item[] Answer: \answerYes{}. 
    \item[] Justification: The work studies temporal leakage and cutoff-aware verification in LLMs using public benchmark data and standard model evaluations. It does not involve deception, human subjects, private data collection, or harmful deployment.

\item {\bf Broader impacts}
    \item[] Question: Does the paper discuss both potential positive societal impacts and negative societal impacts of the work performed?
    \item[] Answer: \answerYes{}. 
    \item[] Justification: The work can improve the reliability of LLMs in time-sensitive settings such as historical analysis, financial backtesting, and scientific trend analysis. A potential negative use is that models may be trained to conceal the use of future knowledge; the paper frames TCFT as a verification method for reducing leakage rather than enabling misuse.
    
\item {\bf Safeguards}
    \item[] Question: Does the paper describe safeguards that have been put in place for responsible release of data or models that have a high risk for misuse (e.g., pre-trained language models, image generators, or scraped datasets)?
    \item[] Answer: \answerNA{}. 
    \item[] Justification: The paper does not release a high-risk pretrained model, image generator, or unsafe scraped dataset. The released assets are code and temporal critique data for evaluating and improving cutoff adherence.

\item {\bf Licenses for existing assets}
    \item[] Question: Are the creators or original owners of assets (e.g., code, data, models), used in the paper, properly credited and are the license and terms of use explicitly mentioned and properly respected?
    \item[] Answer: \answerYes{}. 
    \item[] Justification: The paper cites the existing ExAnte benchmark, Qwen model backbones, and proprietary LLM APIs used in evaluation or critique generation. The corresponding licenses and terms of use are documented in the appendix.

\item {\bf New assets}
    \item[] Question: Are new assets introduced in the paper well documented and is the documentation provided alongside the assets?
    \item[] Answer: \answerYes{}. 
    \item[] Justification: The paper introduces TCFT training data derived from temporal critique generation and releases the code for constructing and evaluating this data. Documentation for the data format, training pipeline, and evaluation scripts is provided in the repository.

\item {\bf Crowdsourcing and research with human subjects}
    \item[] Question: For crowdsourcing experiments and research with human subjects, does the paper include the full text of instructions given to participants and screenshots, if applicable, as well as details about compensation (if any)? 
    \item[] Answer: \answerNA{}. 
    \item[] Justification: The paper does not involve crowdsourcing or research with human subjects. Temporal critiques are generated by teacher models and evaluated on existing benchmark data.

\item {\bf Institutional review board (IRB) approvals or equivalent for research with human subjects}
    \item[] Question: Does the paper describe potential risks incurred by study participants, whether such risks were disclosed to the subjects, and whether Institutional Review Board (IRB) approvals (or an equivalent approval/review based on the requirements of your country or institution) were obtained?
    \item[] Answer: \answerNA{}. 
    \item[] Justification: The paper does not involve human subjects, user studies, or crowdsourced annotation, so IRB approval is not applicable.

\item {\bf Declaration of LLM usage}
    \item[] Question: Does the paper describe the usage of LLMs if it is an important, original, or non-standard component of the core methods in this research? Note that if the LLM is used only for writing, editing, or formatting purposes and does \emph{not} impact the core methodology, scientific rigor, or originality of the research, declaration is not required.
    \item[] Answer: \answerYes{}. 
    \item[] Justification: The paper uses proprietary LLMs for evaluation and teacher-generated temporal critiques, and fine-tunes Qwen2.5-7B-Instruct and Qwen2.5-14B-Instruct as student backbones. LLMs were also used for language polishing.

\end{enumerate}

%% file: nips.bib
@article{valmeekam2023can,
  title={Can large language models really improve by self-critiquing their own plans?},
  author={Valmeekam, Karthik and Marquez, Matthew and Kambhampati, Subbarao},
  journal={arXiv preprint arXiv:2310.08118},
  year={2023}
}

@article{yuan2024free,
  title={Free process rewards without process labels},
  author={Yuan, Lifan and Li, Wendi and Chen, Huayu and Cui, Ganqu and Ding, Ning and Zhang, Kaiyan and Zhou, Bowen and Liu, Zhiyuan and Peng, Hao},
  journal={arXiv preprint arXiv:2412.01981},
  year={2024}
}

@article{ankner2024critique,
  title={Critique-out-loud reward models},
  author={Ankner, Zachary and Paul, Mansheej and Cui, Brandon and Chang, Jonathan D and Ammanabrolu, Prithviraj},
  journal={arXiv preprint arXiv:2408.11791},
  year={2024}
}

@inproceedings{min2023factscore,
  title={Factscore: Fine-grained atomic evaluation of factual precision in long form text generation},
  author={Min, Sewon and Krishna, Kalpesh and Lyu, Xinxi and Lewis, Mike and Yih, Wen-tau and Koh, Pang and Iyyer, Mohit and Zettlemoyer, Luke and Hajishirzi, Hannaneh},
  booktitle={Proceedings of the 2023 Conference on Empirical Methods in Natural Language Processing},
  pages={12076--12100},
  year={2023}
}

@inproceedings{wang2024math,
  title={Math-shepherd: Verify and reinforce llms step-by-step without human annotations},
  author={Wang, Peiyi and Li, Lei and Shao, Zhihong and Xu, Runxin and Dai, Damai and Li, Yifei and Chen, Deli and Wu, Yu and Sui, Zhifang},
  booktitle={Proceedings of the 62nd Annual Meeting of the Association for Computational Linguistics (Volume 1: Long Papers)},
  pages={9426--9439},
  year={2024}
}

@article{xu2025double,
  title={Double-Checker: Enhancing Reasoning of Slow-Thinking LLMs via Self-Critical Fine-Tuning},
  author={Xu, Xin and Chen, Tianhao and Zhang, Fan and Liu, Wanlong and Li, Pengxiang and Jaiswal, Ajay Kumar and Yan, Yuchen and Hu, Jishan and Wang, Yang and Chen, Hao and others},
  journal={arXiv preprint arXiv:2506.21285},
  year={2025}
}

@inproceedings{Reflexion,
  author       = {Noah Shinn and
                  Federico Cassano and
                  Ashwin Gopinath and
                  Karthik Narasimhan and
                  Shunyu Yao},
  title        = {Reflexion: language agents with verbal reinforcement learning},
  booktitle    = {NeurIPS},
  year         = {2023}
}

@article{gou2023critic,
  title={Critic: Large language models can self-correct with tool-interactive critiquing},
  author={Gou, Zhibin and Shao, Zhihong and Gong, Yeyun and Shen, Yelong and Yang, Yujiu and Duan, Nan and Chen, Weizhu},
  journal={arXiv preprint arXiv:2305.11738},
  year={2023}
}

@inproceedings{Self-Refine,
  author       = {Aman Madaan and
                  Niket Tandon and
                  Prakhar Gupta and
                  Skyler Hallinan and
                  Luyu Gao and
                  Sarah Wiegreffe and
                  Uri Alon and
                  Nouha Dziri and
                  Shrimai Prabhumoye and
                  Yiming Yang and
                  Shashank Gupta and
                  Bodhisattwa Prasad Majumder and
                  Katherine Hermann and
                  Sean Welleck and
                  Amir Yazdanbakhsh and
                  Peter Clark},
  title        = {Self-Refine: Iterative Refinement with Self-Feedback},
  booktitle    = {NeurIPS},
  year         = {2023}
}

@inproceedings{Lookaheadbias,
  title={Lookahead bias in pretrained language models},
  author={Sarkar, Suproteem K and Vafa, Keyon},
  booktitle={ICML 2025 Workshop on Reliable and Responsible Foundation Models},
  year={2024}
}

@article{NPO,
  author  = {Ruiqi Zhang and
             Licong Lin and
             Yu Bai and
             Song Mei},
  title   = {Negative Preference Optimization: From Catastrophic Collapse to Effective Unlearning},
  journal = {CoRR},
  volume  = {abs/2404.05868},
  year    = {2024}
}

@article{CritiqueFineTuning,
  author  = {Yubo Wang and
             Xiang Yue and
             Wenhu Chen},
  title   = {{Critique Fine-Tuning}: Learning to Critique is More Effective than Learning to Imitate},
  journal = {CoRR},
  volume  = {abs/2501.17703},
  year    = {2025}
}

@inproceedings{StreamingQA,
  author    = {Adam Liska and
               Tom{\'{a}}s Kocisk{\'{y}} and
               Elena Gribovskaya and
               Tayfun Terzi and
               Eren Sezener and
               Devang Agrawal and
               Cyprien de Masson d'Autume and
               Tim Scholtes and
               Manzil Zaheer and
               Susannah Young and
               Ellen Gilsenan{-}McMahon and
               Sophia Austin and
               Phil Blunsom and
               Angeliki Lazaridou},
  title     = {{StreamingQA}: {A} Benchmark for Adaptation to New Knowledge over Time in Question Answering Models},
  booktitle = {{ICML}},
  series    = {Proceedings of Machine Learning Research},
  pages     = {13604--13622},
  publisher = {{PMLR}},
  year      = {2022}
}

@inproceedings{UnrollingSGD,
  title        = {Unrolling sgd: Understanding factors influencing machine unlearning},
  author       = {Thudi, Anvith and Deza, Gabriel and Chandrasekaran, Varun and Papernot, Nicolas},
  booktitle    = {2022 IEEE 7th European Symposium on Security and Privacy (EuroS\&P)},
  pages        = {303--319},
  year         = {2022},
  organization = {IEEE}
}

@article{ding2026cuse,
  title={CUSE: Fine-Grained Multimodal Unlearning with Counterfactual Guidance and On-Policy Self-Distillation},
  author={Ding, Chenlu and Wu, Jiancan and Luo, Yanchen and Zhang, Tianyu and Yuan, Yancheng and Wang, Xiang and He, Xiangnan},
  journal={arXiv preprint arXiv:2610.14763},
  year={2026}
}

@inproceedings{llmeraser,
  author       = {Chenlu Ding and
                  Jiancan Wu and
                  Yancheng Yuan and
                  Jinda Lu and
                  Kai Zhang and
                  Alex Su and
                  Xiang Wang and
                  Xiangnan He},
  title        = {Unified Parameter-Efficient Unlearning for LLMs},
  booktitle    = {{ICLR}},
  publisher    = {OpenReview.net},
  year         = {2025}
}

@inproceedings{InContextUnlearning,
  author    = {Martin Pawelczyk and
               Seth Neel and
               Himabindu Lakkaraju},
  title     = {In-Context Unlearning: Language Models as Few-Shot Unlearners},
  booktitle = {{ICML}},
  publisher = {OpenReview.net},
  year      = {2024}
}

@inproceedings{TRAM,
  author    = {Yuqing Wang and
               Yun Zhao},
  title     = {{TRAM:} Benchmarking Temporal Reasoning for Large Language Models},
  booktitle = {{ACL} (Findings)},
  series    = {Findings of {ACL}},
  pages     = {6389--6415},
  publisher = {Association for Computational Linguistics},
  year      = {2024}
}

@inproceedings{TimeBench,
  author    = {Zheng Chu and
               Jingchang Chen and
               Qianglong Chen and
               Weijiang Yu and
               Haotian Wang and
               Ming Liu and
               Bing Qin},
  title     = {{TimeBench}: {A} Comprehensive Evaluation of Temporal Reasoning Abilities in Large Language Models},
  booktitle = {{ACL} {(1)}},
  pages     = {1204--1228},
  publisher = {Association for Computational Linguistics},
  year      = {2024}
}

@inproceedings{OrderMatters,
  author       = {Jie Zeng and
                  Qianyu He and
                  Qingyu Ren and
                  Jiaqing Liang and
                  Weikang Zhou and
                  Zeye Sun and
                  Fei Yu and
                  Yanghua Xiao},
  title        = {Order Matters: Investigate the Position Bias in Multi-constraint Instruction
                  Following},
  booktitle    = {{ACL} (Findings)},
  series       = {Findings of {ACL}},
  volume       = {{ACL} 2025},
  pages        = {12479--12492},
  publisher    = {Association for Computational Linguistics},
  year         = {2025}
}

@inproceedings{TestOfTime,
  author    = {Bahare Fatemi and
               Mehran Kazemi and
               Anton Tsitsulin and
               Karishma Malkan and
               Jinyeong Yim and
               John Palowitch and
               Sungyong Seo and
               Jonathan Halcrow and
               Bryan Perozzi},
  title     = {{Test of Time}: {A} Benchmark for Evaluating LLMs on Temporal Reasoning},
  booktitle = {{ICLR}},
  publisher = {OpenReview.net},
  year      = {2025}
}

@inproceedings{ExAnte,
  author    = {Yachuan Liu and
               Xiaochun Wei and
               Lin Shi and
               Xinnuo Li and
               Bohan Zhang and
               Paramveer Dhillon and
               Qiaozhu Mei},
  title     = {{ExAnte}: A Benchmark for Ex-Ante Inference in Large Language Models},
  booktitle = {{EACL} (Volume 1: Long Papers)},
  pages     = {1551--1571},
  publisher = {Association for Computational Linguistics},
  year      = {2026}
}

@article{gruver2023large,
  title={Large language models are zero-shot time series forecasters},
  author={Gruver, Nate and Finzi, Marc and Qiu, Shikai and Wilson, Andrew G},
  journal={Advances in neural information processing systems},
  volume={36},
  pages={19622--19635},
  year={2023}
}

@misc{Anthropic2026ClaudeOpus46,
  author       = {{Anthropic}},
  title        = {Introducing Claude Opus 4.6},
  year         = {2026},
  howpublished = {\url{https://www.anthropic.com/news/claude-opus-4-6}},
  note         = {Accessed: 2026-05-07}
}

@misc{Google2026Gemini31ProPreview,
  author       = {{Google}},
  title        = {Gemini 3.1 Pro Preview},
  year         = {2026},
  howpublished = {\url{https://ai.google.dev/gemini-api/docs/models/gemini-3.1-pro-preview}},
  note         = {Accessed: 2026-05-07}
}

@misc{OpenAI2026GPT54,
  title        = {Introducing GPT-5.4},
  author       = {{OpenAI}},
  year         = {2026},
  howpublished = {\url{https://openai.com/index/introducing-gpt-5-4/}},
  note         = {Accessed: 2026-05-07}
}

@misc{Wikipedia,
  author       = {{Wikimedia Foundation}},
  title        = {Wikipedia},
  year         = {2026},
  howpublished = {\url{https://www.wikipedia.org/}},
  note         = {Accessed: 2026-05-07}
}

@article{brown2020language,
  title={Language models are few-shot learners},
  author={Brown, Tom and Mann, Benjamin and Ryder, Nick and Subbiah, Melanie and Kaplan, Jared D and Dhariwal, Prafulla and Neelakantan, Arvind and Shyam, Pranav and Sastry, Girish and Askell, Amanda and others},
  journal={Advances in neural information processing systems},
  volume={33},
  pages={1877--1901},
  year={2020}
}

@article{jin2023time,
  title={Time-llm: Time series forecasting by reprogramming large language models},
  author={Jin, Ming and Wang, Shiyu and Ma, Lintao and Chu, Zhixuan and Zhang, James Y and Shi, Xiaoming and Chen, Pin-Yu and Liang, Yuxuan and Li, Yuan-Fang and Pan, Shirui and others},
  journal={arXiv preprint arXiv:2310.01728},
  year={2023}
}

@article{FinLLMs,
  author       = {Jean Lee and
                  Nicholas Stevens and
                  Soyeon Caren Han},
  title        = {Large Language Models in Finance (FinLLMs)},
  journal      = {Neural Comput. Appl.},
  volume       = {37},
  number       = {30},
  pages        = {24853--24867},
  year         = {2025}
}

@article{Accounting,
  author       = {Ties de Kok},
  title        = {ChatGPT for Textual Analysis? How to Use Generative LLMs in Accounting
                  Research},
  journal      = {Manag. Sci.},
  volume       = {71},
  number       = {9},
  pages        = {7888--7906},
  year         = {2025}
}

@inproceedings{history,
  author       = {Shuhang Lin and
                  Wenyue Hua and
                  Lingyao Li and
                  Che{-}Jui Chang and
                  Lizhou Fan and
                  Jianchao Ji and
                  Hang Hua and
                  Mingyu Jin and
                  Jiebo Luo and
                  Yongfeng Zhang},
  title        = {BattleAgent: Multi-modal Dynamic Emulation on Historical Battles to
                  Complement Historical Analysis},
  booktitle    = {{EMNLP} (System Demonstrations)},
  pages        = {172--181},
  publisher    = {Association for Computational Linguistics},
  year         = {2024}
}

@inproceedings{science,
  author       = {Jinheon Baek and
                  Sujay Kumar Jauhar and
                  Silviu Cucerzan and
                  Sung Ju Hwang},
  title        = {ResearchAgent: Iterative Research Idea Generation over Scientific
                  Literature with Large Language Models},
  booktitle    = {{NAACL} (Long Papers)},
  pages        = {6709--6738},
  publisher    = {Association for Computational Linguistics},
  year         = {2025}
}

@article{GPT-4,
  author       = {OpenAI},
  title        = {{GPT-4} Technical Report},
  journal      = {CoRR},
  volume       = {abs/2303.08774},
  year         = {2023}
}

@article{Gemini,
  author       = {Gemini Team},
  title        = {Gemini: {A} Family of Highly Capable Multimodal Models},
  journal      = {CoRR},
  volume       = {abs/2312.11805},
  year         = {2023}
}

@article{Dated_Data,
  author       = {Jeffrey Cheng and
                  Marc Marone and
                  Orion Weller and
                  Dawn J. Lawrie and
                  Daniel Khashabi and
                  Benjamin Van Durme},
  title        = {Dated Data: Tracing Knowledge Cutoffs in Large Language Models},
  journal      = {CoRR},
  volume       = {abs/2403.12958},
  year         = {2024}
}

@inproceedings{cot,
  author       = {Jason Wei and
                  Xuezhi Wang and
                  Dale Schuurmans and
                  Maarten Bosma and
                  Brian Ichter and
                  Fei Xia and
                  Ed H. Chi and
                  Quoc V. Le and
                  Denny Zhou},
  title        = {Chain-of-Thought Prompting Elicits Reasoning in Large Language Models},
  booktitle    = {NeurIPS},
  year         = {2022}
}

@inproceedings{self-verification,
  author       = {Ziwei Ji and
                  Tiezheng Yu and
                  Yan Xu and
                  Nayeon Lee and
                  Etsuko Ishii and
                  Pascale Fung},
  title        = {Towards Mitigating {LLM} Hallucination via Self Reflection},
  booktitle    = {{EMNLP} (Findings)},
  series       = {Findings of {ACL}},
  pages        = {1827--1843},
  publisher    = {Association for Computational Linguistics},
  year         = {2023}
}

@inproceedings{lora,
  author       = {Edward J. Hu and
                  Yelong Shen and
                  Phillip Wallis and
                  Zeyuan Allen{-}Zhu and
                  Yuanzhi Li and
                  Shean Wang and
                  Lu Wang and
                  Weizhu Chen},
  title        = {LoRA: Low-Rank Adaptation of Large Language Models},
  booktitle    = {{ICLR}},
  publisher    = {OpenReview.net},
  year         = {2022}
}

@inproceedings{sft,
  author       = {Long Ouyang and
                  Jeffrey Wu and
                  Xu Jiang and
                  Diogo Almeida and
                  Carroll L. Wainwright and
                  Pamela Mishkin and
                  Chong Zhang and
                  Sandhini Agarwal and
                  Katarina Slama and
                  Alex Ray and
                  John Schulman and
                  Jacob Hilton and
                  Fraser Kelton and
                  Luke Miller and
                  Maddie Simens and
                  Amanda Askell and
                  Peter Welinder and
                  Paul F. Christiano and
                  Jan Leike and
                  Ryan Lowe},
  title        = {Training language models to follow instructions with human feedback},
  booktitle    = {NeurIPS},
  year         = {2022}
}

@inproceedings{ft,
  author       = {Jason Wei and
                  Maarten Bosma and
                  Vincent Y. Zhao and
                  Kelvin Guu and
                  Adams Wei Yu and
                  Brian Lester and
                  Nan Du and
                  Andrew M. Dai and
                  Quoc V. Le},
  title        = {Finetuned Language Models are Zero-Shot Learners},
  booktitle    = {{ICLR}},
  publisher    = {OpenReview.net},
  year         = {2022}
}

@inproceedings{instrcut,
  author       = {Victor Sanh and
                  Albert Webson and
                  Colin Raffel and
                  Stephen H. Bach and
                  Lintang Sutawika and
                  Zaid Alyafeai and
                  Antoine Chaffin and
                  Arnaud Stiegler and
                  Arun Raja and
                  Manan Dey and
                  M Saiful Bari and
                  Canwen Xu and
                  Urmish Thakker and
                  Shanya Sharma Sharma and
                  Eliza Szczechla and
                  Taewoon Kim and
                  Gunjan Chhablani and
                  Nihal V. Nayak and
                  Debajyoti Datta and
                  Jonathan Chang and
                  Mike Tian{-}Jian Jiang and
                  Han Wang and
                  Matteo Manica and
                  Sheng Shen and
                  Zheng Xin Yong and
                  Harshit Pandey and
                  Rachel Bawden and
                  Thomas Wang and
                  Trishala Neeraj and
                  Jos Rozen and
                  Abheesht Sharma and
                  Andrea Santilli and
                  Thibault F{\'{e}}vry and
                  Jason Alan Fries and
                  Ryan Teehan and
                  Teven Le Scao and
                  Stella Biderman and
                  Leo Gao and
                  Thomas Wolf and
                  Alexander M. Rush},
  title        = {Multitask Prompted Training Enables Zero-Shot Task Generalization},
  booktitle    = {{ICLR}},
  publisher    = {OpenReview.net},
  year         = {2022}
}

@inproceedings{Self-Instruct,
  author       = {Yizhong Wang and
                  Yeganeh Kordi and
                  Swaroop Mishra and
                  Alisa Liu and
                  Noah A. Smith and
                  Daniel Khashabi and
                  Hannaneh Hajishirzi},
  title        = {Self-Instruct: Aligning Language Models with Self-Generated Instructions},
  booktitle    = {{ACL} {(1)}},
  pages        = {13484--13508},
  publisher    = {Association for Computational Linguistics},
  year         = {2023}
}

@inproceedings{Verify_step,
  author       = {Hunter Lightman and
                  Vineet Kosaraju and
                  Yuri Burda and
                  Harrison Edwards and
                  Bowen Baker and
                  Teddy Lee and
                  Jan Leike and
                  John Schulman and
                  Ilya Sutskever and
                  Karl Cobbe},
  title        = {Let's Verify Step by Step},
  booktitle    = {{ICLR}},
  publisher    = {OpenReview.net},
  year         = {2024}
}

@article{judgement,
  author       = {Jonathan Uesato and
                  Nate Kushman and
                  Ramana Kumar and
                  H. Francis Song and
                  Noah Y. Siegel and
                  Lisa Wang and
                  Antonia Creswell and
                  Geoffrey Irving and
                  Irina Higgins},
  title        = {Solving math word problems with process- and outcome-based feedback},
  journal      = {CoRR},
  volume       = {abs/2211.14275},
  year         = {2022}
}

@article{qwen-2.5,
  author       = {An Yang and
                  Baosong Yang and
                  Beichen Zhang and
                  Binyuan Hui and
                  Bo Zheng and
                  Bowen Yu and
                  Chengyuan Li and
                  Dayiheng Liu and
                  Fei Huang and
                  Haoran Wei and
                  Huan Lin and
                  Jian Yang and
                  Jianhong Tu and
                  Jianwei Zhang and
                  Jianxin Yang and
                  Jiaxi Yang and
                  Jingren Zhou and
                  Junyang Lin and
                  Kai Dang and
                  Keming Lu and
                  Keqin Bao and
                  Kexin Yang and
                  Le Yu and
                  Mei Li and
                  Mingfeng Xue and
                  Pei Zhang and
                  Qin Zhu and
                  Rui Men and
                  Runji Lin and
                  Tianhao Li and
                  Tingyu Xia and
                  Xingzhang Ren and
                  Xuancheng Ren and
                  Yang Fan and
                  Yang Su and
                  Yichang Zhang and
                  Yu Wan and
                  Yuqiong Liu and
                  Zeyu Cui and
                  Zhenru Zhang and
                  Zihan Qiu},
  title        = {Qwen2.5 Technical Report},
  journal      = {CoRR},
  volume       = {abs/2412.15115},
  year         = {2024}
}

@inproceedings{temporal1,
  author       = {Zhaochen Su and
                  Juntao Li and
                  Jun Zhang and
                  Tong Zhu and
                  Xiaoye Qu and
                  Pan Zhou and
                  Yan Bowen and
                  Yu Cheng and
                  Min Zhang},
  title        = {Living in the Moment: Can Large Language Models Grasp Co-Temporal
                  Reasoning?},
  booktitle    = {{ACL} {(1)}},
  pages        = {13014--13033},
  publisher    = {Association for Computational Linguistics},
  year         = {2024}
}

@inproceedings{temporal2,
  author       = {Qingyu Tan and
                  Hwee Tou Ng and
                  Lidong Bing},
  title        = {Towards Benchmarking and Improving the Temporal Reasoning Capability
                  of Large Language Models},
  booktitle    = {{ACL} {(1)}},
  pages        = {14820--14835},
  publisher    = {Association for Computational Linguistics},
  year         = {2023}
}

@inproceedings{time1,
  author       = {Shahriar Golchin and
                  Mihai Surdeanu},
  title        = {Time Travel in LLMs: Tracing Data Contamination in Large Language
                  Models},
  booktitle    = {{ICLR}},
  publisher    = {OpenReview.net},
  year         = {2024}
}

@inproceedings{time2,
  author       = {Manley Roberts and
                  Himanshu Thakur and
                  Christine Herlihy and
                  Colin White and
                  Samuel Dooley},
  title        = {To the Cutoff... and Beyond? {A} Longitudinal Perspective on {LLM}
                  Data Contamination},
  booktitle    = {{ICLR}},
  publisher    = {OpenReview.net},
  year         = {2024}
}

@article{unlearningsurvey,
  author       = {Zheyuan Liu and
                  Guangyao Dou and
                  Zhaoxuan Tan and
                  Yijun Tian and
                  Meng Jiang},
  title        = {Machine Unlearning in Generative {AI:} {A} Survey},
  journal      = {CoRR},
  volume       = {abs/2407.20516},
  year         = {2024}
}

@inproceedings{task,
  author       = {Tianyu Zhang and
                  Chenlu Ding and
                  Junfeng Fang and
                  Jiancan Wu and
                  Xiang Wang},
  title        = {Task-Aligned Unlearning for Multimodal Large Language Models},
  booktitle    = {{WWW} (Companion Volume)},
  pages        = {1386--1390},
  publisher    = {{ACM}},
  year         = {2026}
}

@article{mllmeraser,
  author       = {Chenlu Ding and
                  Jiancan Wu and
                  Leheng Sheng and
                  Fan Zhang and
                  Yancheng Yuan and
                  Xiang Wang and
                  Xiangnan He},
  title        = {MLLMEraser: Achieving Test-Time Unlearning in Multimodal Large Language
                  Models through Activation Steering},
  journal      = {CoRR},
  volume       = {abs/2510.04217},
  year         = {2025}
}

@inproceedings{SaferUnlearning,
  author       = {Zheyuan Liu and
                  Guangyao Dou and
                  Zhaoxuan Tan and
                  Yijun Tian and
                  Meng Jiang},
  title        = {Towards Safer Large Language Models through Machine Unlearning},
  booktitle    = {{ACL} (Findings)},
  series       = {Findings of {ACL}},
  pages        = {1817--1829},
  publisher    = {Association for Computational Linguistics},
  year         = {2024}
}
